# Functional Classification of Spiking Signal Data Using Artificial Intelligence Techniques: A Review

Danial Sharifrazi[1*], Nouman Javed[1,2], Javad Hassannataj Joloudari[3,4,5], Roohallah Alizadehsani[1], Saadat Behzadi[6], Prasad N. Paradkar[2], Ru-San Tan[7,8], U. Rajendra Acharya[9,10], and Asim Bhatti[1]

*[1]Institute for Intelligent Systems Research and Innovations (IISRI), Deakin University, Geelong, Australia*
*[2]CSIRO Health and Biosecurity, Australian Animal Health Laboratory, Geelong, Australia*
*[3]Department of Computer Engineering, Technical and Vocational University (TVU), Tehran, Iran*
*[4]Faculty of Electrical and Computer Engineering, University of Birjand, Birjand, Iran*
*[5]Department of Computer Engineering, Islamic Azad University, Babol, Iran*
*[6]Department of Electronic Engineering, University of Bologna, Bologna, Italy*
*[7]Department of Cardiology, National Heart Centre Singapore, Singapore, Singapore 169609*
*[8]Duke-NUS Medical School, Singapore, Singapore 169857*
*[9]School of Mathematics, Physics and Computing, University of Southern Queensland, Springfield, Australia*
*[10]Center for Health Research, University of Southern Queensland, Springfield, Australia*

## Abstract

**Background:** Human brain neuron activities are incredibly significant nowadays. Neuronal behavior is assessed by analyzing signal data such as extracellular recording, which can offer scientists valuable information about diseases and neuron activities. One of the difficulties researchers confront while evaluating these signals is the existence of large volumes of spike data. Spikes are significant components of signal data that can happen as a consequence of vital biomarkers or physical issues such as electrode movements. Hence, distinguishing types of spikes is essential. From this spot, the spike classification concept commences. Previously, researchers classified spikes manually. The manual classification was not precise enough, as it involves extensive analysis. Consequently, Artificial Intelligence (AI) was introduced into neuroscience to assist clinicians in classifying spikes correctly.

**Objectives:** Recognizing noises from spikes produced by neural activity causes the spike classification task to bear a significant demand. Classifying spikes accurately and quickly reveals the role of AI in the scope of spike classification. This review provides an in-depth discussion of the importance and use of AI in spike classification. This work organizes materials in the spike classification field for future studies and fully describes how spikes are recognized. Therefore, the existing datasets are described first. The topic of spike classification is then separated into three major components: preprocessing, classification, and evaluation.

[*] Corresponding author at: Institute for Intelligent Systems Research and Innovations (IISRI), Deakin University, Geelong, Victoria 3216, Australia

Email Address: s222619018@deakin.edu.au (D. Sharifrazi)

Each of these sections introduces existing methods and determines their importance. Having been summarized and compared, more efficient algorithms are highlighted. The primary goal of this work is to provide a perspective on spike classification for future research, as well as a thorough grasp of the methodologies and issues involved.

**Methodology:** In this work, numerous studies were extracted from various databases. The PRISMA-related research guidelines were then used to choose papers. Then, research studies based on spike classification using machine learning and deep learning approaches with effective preprocessing were selected. Although there are research papers on spike sorting using the keyword spike, the primary focus of this study is on spike classification.

**Results:** Finally, 47 papers were selected for in-depth review. First, useful information on the datasets for these papers is supplied. In addition, preprocessing approaches, classification methods, and ultimate performance are investigated in each of these studies. The material is then summarized. Furthermore, the fundamental concerns regarding spike classification raised in the opening of this paper are thoroughly addressed throughout the review. Our reviewing outcomes illustrate that support vector machine and clustering-based algorithms drastically influence machine learning methods in terms of high accuracy and many uses. Moreover, convolutional neural networks, spiky neural networks, and attention-based techniques can classify spikes with considerable functionality among deep learning methods.

**Conclusion:** Various preprocessing and classification techniques have been used practically to classify extracted signal data from patients in medical institutions. Our review emphasizes the importance of classifying spikes in neuroscience applications with machine learning and deep learning models. This can provide precious insights and hands-on solutions for using AI to classify real-world medical data.

## 1. Introduction

In the brain, neurons initiate and propagate electrical signals among themselves that serve as communication across the various brain areas [1]. These neuronal activities, which were initially observed on invasive extracellular microelectrodes [2], demonstrate discrete waveform shapes ("spikes") that can be influenced by various factors, e.g., neuron shape and membrane properties, microenvironment, and distance from the recording electrode. Indeed, the spike patterns of recorded neuronal electrical activity offer a window into understanding brain functioning in health and disease. As neuroscience research grew, efficient and precise analysis of spike patterns has become an essential tool.

Spike pattern analysis encompasses spike identification, as well as spike feature extraction and classification [3]. Spike identification seeks to identify potential brain spikes in samples recorded from either invasive microelectrodes or noninvasive scalp surface electroencephalography (EEG) electrodes. An electrode may detect several spikes near a large number of neurons. Signal preprocessing using techniques like filtering [4] is performed to extract waveforms induced by brain activity from those generated by random voltage fluctuations and noise, which may be attributable to electrode decoding or signal voltages near the electrode tips (Figure 1). Spike classification, involving methodology for estimating spike rates, synchrony analysis, and neural coding, is the process of grouping brain spikes with similar qualities into individual clusters, i.e., the spikes are considered to belong to the same isolated unit.

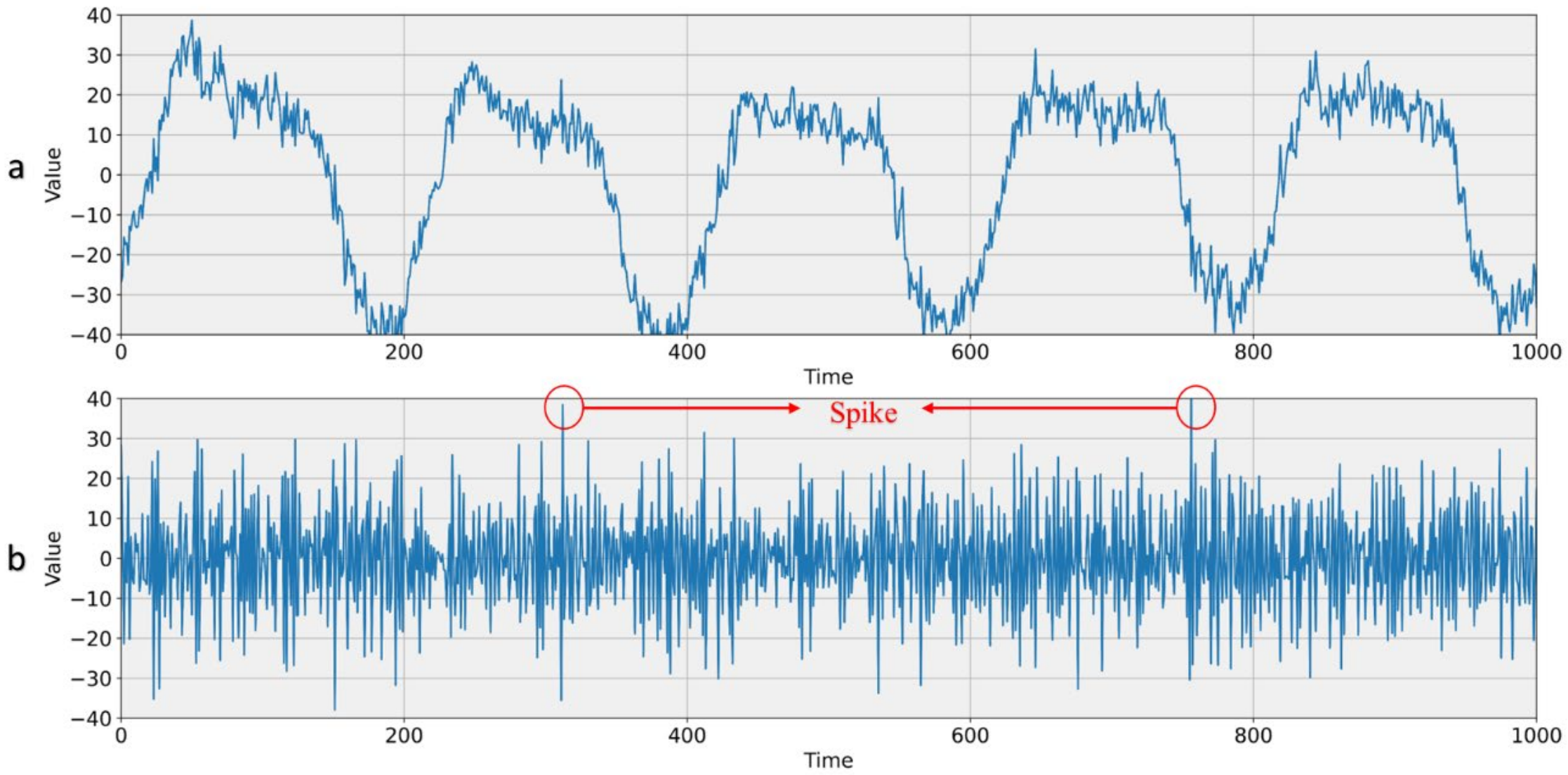

Figure 1. a) Raw extracellular recording signal. b) After preprocessing, spikes are recovered for downstream analysis.

The effect of spike classification on the robustness of the neuroscience study findings cannot be denied [5]. Accurate classification model development requires knowledge of signals, signal processing and data mining techniques [6]. Traditionally, neuron data are manually sorted into true neuron activity and noise based on the waveform; however, this process is subject to observer expertise and human biases. Spike classification can be based on various specified objective spike characteristics. Lemon et al. classified spikes based on shape analysis [7]; and Balaram et al, spatiotemporal mapping and electrical source analysis [8].

In recent years, AI has revolutionized the medical field, significantly advancing both behavioral [9-12] and neural research [13]. AI approaches, especially deep learning [14], are being adopted for spike classification [15], e.g., in disease states, like seizures [8, 16, 17]. Modeled after neurons in the human brain [18]. Deep learning uses non-linear processing units ("neurons") to execute complex calculations and extract signal features automatically. Outperforming classic machine learning methods in signal analysis performance, deep learning demands comparatively heavier computational resources [19], and a large quantity of data and several iterations for training, which may render these methods less feasible for wearable devices [18, 19]. Despite current limitations, AI holds potential for spike pattern analysis application to brain-computer interfaces (BCI), e.g., mobile EEG-enabled devices can be harnessed for communications or motor rehabilitation needs in patients who have lost speech or movement capabilities [20].

In this work, we review AI techniques in spike pattern analysis. AI techniques have been employed for data preprocessing and classification [21]. In the data preprocessing phase, Veerabhadrappa et al. have comprehensively surveyed spike data clustering for data preprocessing [2]. Accordingly, while the current study will also discuss pertinent data preprocessing approaches, our primary focus is on spike classification, which we have stratified using machine learning and deep learning methods. Figure 2 shows the targeted algorithms briefly.

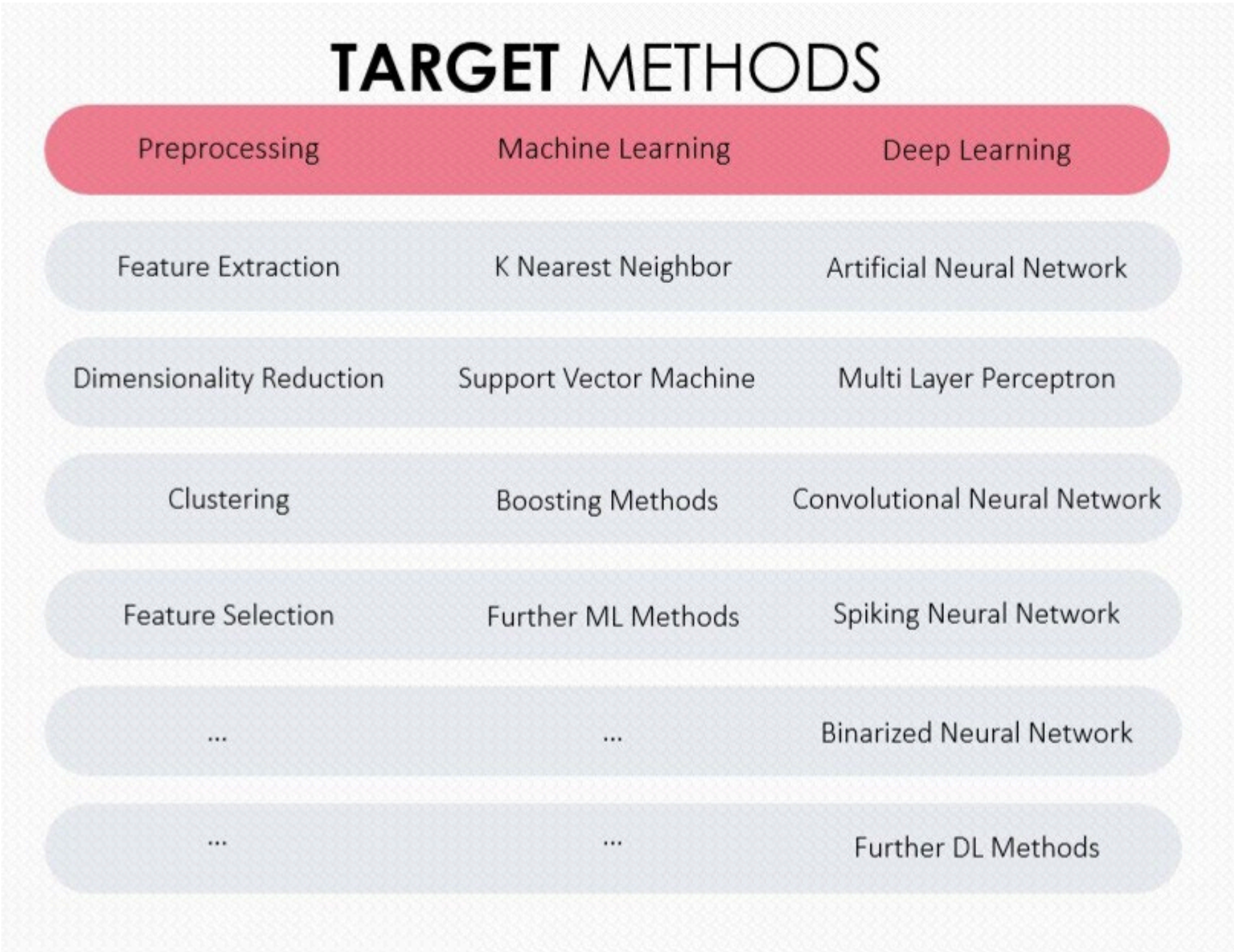

Figure 2. Algorithms employed in preprocessing, machine learning, and deep learning methods in this review.

### *1.1 Related review papers*

Recent review articles have been published on spiking neural networks, but data preprocessing and classification with other machine learning methods were lacking [22-25]. Pietrzak et al. reviewed methods related to spiking neural networks, focusing on their computational complexity. However, AI methods in the field of signal classification were not addressed [26]. Shoeibi et al. reviewed deep learning methods for signal data analysis in seizure disease, but did not consider machine learning methods and studies related to spike classification [6]. Abd El-Samie et al. surveyed spike classification methods on EEG and magnetoencephalography (MEG) data, but examined neither machine learning methods nor AI-based preprocessing [27]. Xu and Xia reviewed feature extraction and deep learning methods used in EEG signal classification, but did not consider other preprocessing and machine learning methods [28]. Parsa et al. reviewed deep learning methods for EEG-based signal and image data classification, but did not study spike classification and AI-based methods related to preprocessing and classification [29]. Auge et al. reviewed encoding methods in signal processing and spiking neural network methods, but did not provide insight into preprocessing-related machine learning methods and signal classification-related deep learning methods [30]. Table 1 compares existing review papers with our proposed study. It may be noted that our work focuses on automated neural spike classification using machine learning and deep learning techniques, using multi-dimensional signals.

Table 1. Summary of related review papers published in 2019 or later.

| *Study* | *Title* | *Scope* | *Number of reviewed papers* | *Key Contributions* | *Limitations* |
|---|---|---|---|---|---|
| Shoeibi et al., 2021 [6] | Epileptic Seizure Detection Using Deep Learning Techniques: A Review | Epileptic seizure detection using EEG and MRI data | 94 | Covers DL architectures for EEG and MRI | Excludes machine learning methods; not focused on extracellular spike data |
| Xu et al., 2023 [28] | EEG Signal Classification and Feature Extraction Methods Based on Deep Learning: A Review | Application of deep learning algorithms on EEG data | 7 | Compares DL and feature extraction methods for EEG classification | Not focused on extracellular spikes; excludes ML techniques |
| Parsa et al., 2023 [29] | EEG-based classification of individuals with neuropsychiatric disorders using deep neural networks: A systematic review of current status and future directions | Neuropsychiatric disorders classification | 82 | Compares DL architectures for EEG classification | Excludes ML and preprocessing techniques |
| Zhang et al., 2023 [31] | Spike sorting algorithms and their efficient hardware implementation: a comprehensive survey | Spike sorting | N/A | Covers DL architectures for spike sorting algorithms and relevant hardware and software packages | Excludes ML techniques |
| Meyer et al., 2024 [32] | Deep learning-based spike sorting: a survey | Spike sorting | 24 | Covers DL and feature extraction methods for spike sorting | Excludes ML and preprocessing techniques |
| Ganguly et al., 2024 [33] | Spike frequency adaptation: bridging neural models and neuromorphic applications | Spike frequency adaptation | 12 | Covers adaptive neuron models | Excludes ML techniques; not provide detailed comparison |
| Quan et al., 2024 [34] | A comprehensive review of spike sorting algorithms in neuroscience | Spike sorting | N/A | Covers DL, clustering, and feature extraction methods for spike sorting | Excludes ML techniques |

| Current work | Functional classification of spiking signal data using artificial intelligence techniques: a review | Spike classification of extracellular data | 47 | Comprehensive review of preprocessing, ML, and DL approaches on extracellular data; compares models, features, datasets, and evaluation metrics | Focused on preprocessing, ML, and DL methods for extracellular data classification |
|---|---|---|---|---|---|

Table 1 shows a summary of recent review papers published in the field of spike or signal classification based on AI methods. There are some limitations in the publications that have not been addressed so far, left out as research gaps in the scope of spike classification review papers. These research gaps motivated us to address the limitations of previous work more closely. As examples of the limitations of previous review studies, it can be mentioned that the reviews of EEG and MEG data types have been considered so far. No review paper has been published that thoroughly examines the AI methods implemented on extracellular data. Also, all previous surveys have reviewed deep learning or machine learning methods or methods for extracting features separately for signal data and no publication has reviewed and compared all methods together. Therefore, a survey that carefully examines all preprocessing methods, all traditional machine learning methods, and new deep learning methods has not been published. This issue motivated us to explore the role of all preprocessing, machine learning, and deep learning methods in the field of spike classification in this survey.

This work fills the gaps in the realm of spike classification with a broader perspective. In this way, it analyzes and examines in detail the paper presented with conventional preprocessing methods for spike classification, including feature extraction, dimensionality reduction, clustering, and feature selection methods. Also, in the section of classification by AI models, it not only discusses and examines traditional machine learning models, but also mentions more modern technologies, including all deep learning methods, as well as ensemble and attention-based models. This article also carefully analyzes the trend of using the aforementioned models since 1994 and the trend and extent of use of each model over time. Our motivation for presenting this systematic survey was to provide researchers with a coherent view of the field of spike classification with extracellular data through AI-based techniques. We also collected a platform for in-depth understanding of spike classification and its evolution, enabling researchers to identify gaps, challenges, and potential prospects in this field and draw a broader vision for their future work.

### *1.2 Objectives and Overview*

“This review seeks to answer the following questions:

- What are the datasets available for spike classification”
- What are the different preprocessing techniques used for spike classification?

• How do machine learning versus deep learning approaches differ in their performance results? and what are the evaluation metrics used for comparison?

In addition we will discuss current challenges and limitations in AI-based spike classification."

The paper is organized as follows. The first part discusses crucial features in a spiky dataset and how to deal with them to generate useful features. Then, within this framework, machine learning and deep learning algorithms are introduced. Finally, the results and metrics for evaluating the proposed models are given. Thus, the studies will be reviewed in three aspects: preprocessing, classification, and evaluation (Figure 3).

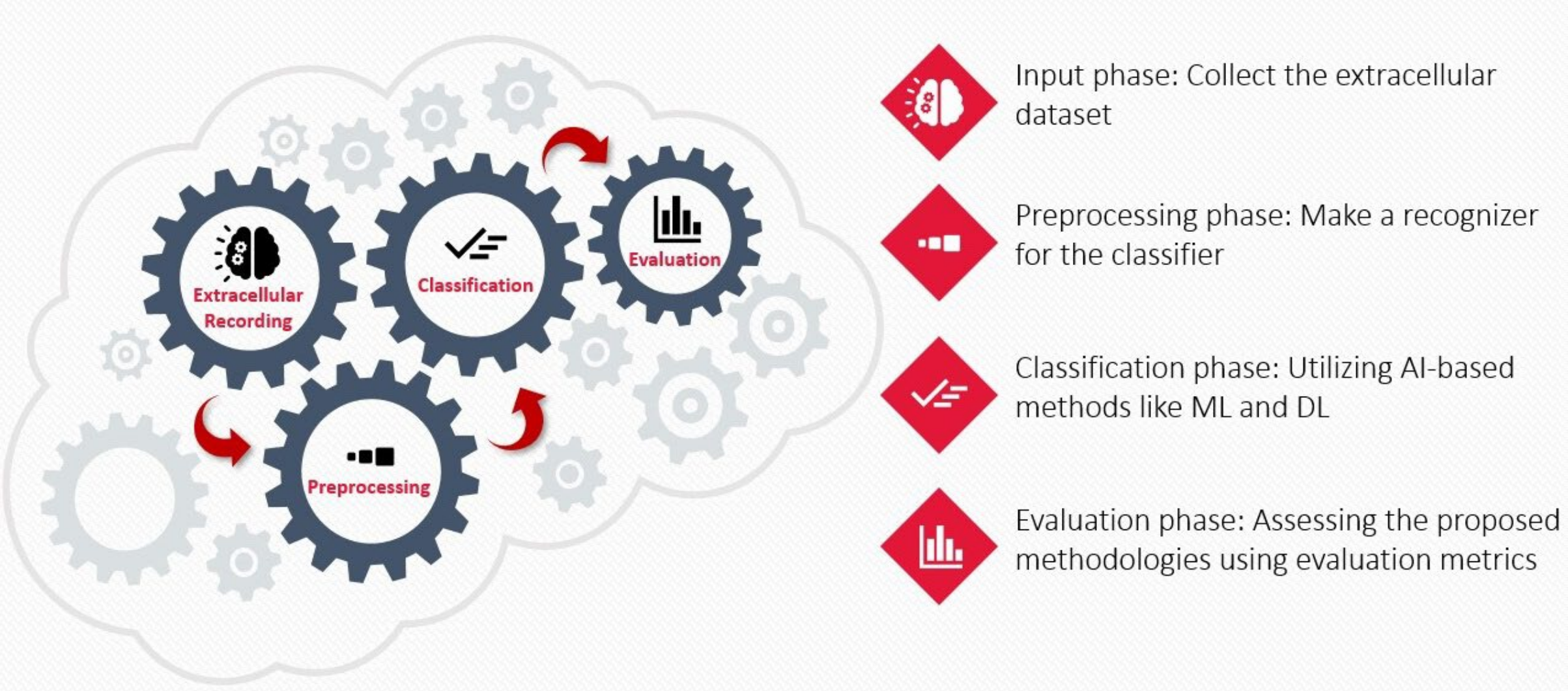

Figure 3. Summary of proposed procedures employed in the related research paper.

This review seeks to answer the following questions: What datasets for spike classification are available, and what are their attributes?

1. What preprocessing methods are most frequently used in extracellular spike classification? How do preprocessing methods influence performance?
2. Which machine learning and deep learning methods are commonly used, and what are the performance results?
3. How do classical machine learning methods compare with deep learning methods for spike classification in terms of accuracy?

The rest of the paper is organized as follows. Section 2 details the search strategy; Section 3 presents and compares the datasets used in related research papers, structured into two focused groups of human studies and high-channel in vivo studies, with the complete dataset table provided in Appendix 3; Section 4 describes the process of spike pattern analysis, including preprocessing and classification; Section 5 provides the results of reviewed studies as a discussion; Section 6 presents main challenges and insights; and Section 7 provides the conclusion and future works. In addition, there are *three* appendices at the end of the article: Appendix 1 provides a list of abbreviations used in the article, and Appendix 2 offers a summary of selected relevant research papers reviewed in this article, and Appendix 3 contains anoverview of datasets used for spike classification.

## 2. Systematic review methodology

### *2.1 Search strategy*

Based on the Preferred Reporting Items for Systematic Reviews and Meta-Analyses (PRISMA) principles [35], we searched on the publisher databases including the Institute of Electrical and Electronics Engineers (IEEE), Elsevier, Institute of Physics publishing (IOP), Springer, Association for Computing Machinery (ACM), Multidisciplinary Digital Publishing Institute (MDPI), Arxiv, and Google Scholar database for publications related to spike pattern analysis using the keywords "spike classification", "spike detection", "machine learning" and "deep learning". We included only studies that used extracellular recording data. We performed duplicate removal using EndNote reference manager, which identified duplicates based on title, DOI, author list, and publication year. After removing duplicates, 189 unique papers remained. From these, 83 non-relevant papers were screened after title and abstract review since they did not utilize AI-based methodologies, leaving 106 publications. The 106 remaining documents were subjected to full-text review, and 59 studies were eliminated because they did not use supervised learning, leaving 47 studies in the last step of our search strategy. Figure 4 illustrates the search strategy succinctly. The summaries of these 47 studies are outlined in Appendix 2.

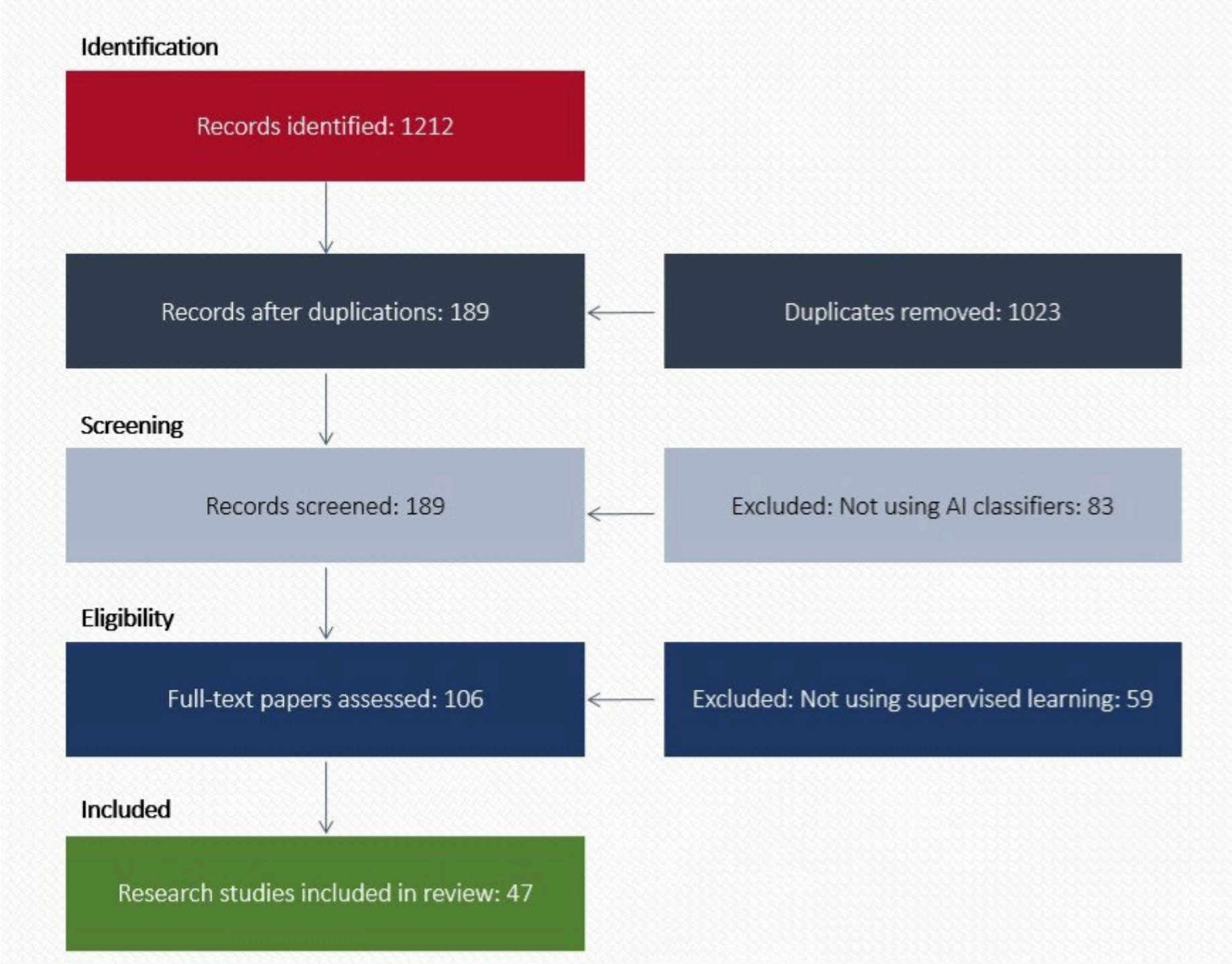

Figure 4. PRISMA flow diagram used to select research papers.

## 3. Overview of available datasets for spike classification

Extracellular recordings provide a widely used approach for monitoring neuronal activity by capturing electrical signals outside the cell membrane. This method allows researchers to track patterns of action potentials generated by neurons without penetrating the cell, making it particularly useful for studying both individual neuron behavior and large-scale neural dynamics [36].

One important application of extracellular recordings is spike classification, a process aimed at identifying which neuron or neuron type generated each recorded spike. Several studies have explored this task using diverse datasets [2, 3], each designed with specific goals and experimental conditions in mind.

These datasets vary significantly in terms of their origin (e.g., in vivo, in vitro, or simulated), biological source (e.g., mouse, monkey, human), and data acquisition methods. Among the most frequently used recording tools are Multi-Electrode Arrays (MEAs), which allow simultaneous recordings from multiple neurons and are employed in both live animal experiments and cultured neuron studies. Other tools include Tetrodes and Utah arrays, particularly in datasets involving higher-resolution or clinical applications.

In addition to differences in hardware, the datasets differ in their sampling rates, the number of recording channels, and spike class diversity. Some datasets were created to evaluate algorithm robustness to noise or signal-to-noise ratio (SNR) variations, while others focus on deep learning-based classification, feature extraction, or benchmark testing under controlled simulation.

A complete list of studies and the dataset attributes is available in Appendix 3. Table 2 summarizes the most relevsant studies that involved human or non-human primate subjects, or in vivo animal experiments with substantial channel counts, and had reported clear performance results.

Table 2. Summary of relevant studies involving human or non-human primate subjects, or in vivo animal experiments with substantial channel counts.

| *Authors* | *Data Source* | *Recording Method* | *Channels* | *Classes* | *Sampling Rate (Hz)* | *Notes* | *Performance Rate* |
|---|---|---|---|---|---|---|---|
| Reddy et al. [37] | Human neurons | Neural Signal Processor | 32 (FPGA) | Multiple | Varied | Real-time classification in a clinical setting | 92.56% Accuracy |
| Valencia and Alimohammad [38] | Human | BNN-based sorting | MEA | 3 | 24,000 | Human cortex recordings for spike sorting | 93% Accuracy |
| Okreghe et al. [39] | Human, non-human primates | CNN-based artifact removal | Varied | Varied | Varied | Focus on artifact correction before classification | 91.53% Accuracy |
| Saif-ur-Rehman et al. [40] | Tetraplegic patients | Self-organizing sorting | 3,072 | Adaptive | Varied | High-density patient data for adaptive sorting | 93% Accuracy |
| Saif-ur-Rehman et al. [41] | Human, non- | Utah array, Micro-wires | 192 | 3 | 24,000 | Cross-species study using | 92.3% Accuracy |

|  | human primates |  |  |  |  | high-density probes |  |
|---|---|---|---|---|---|---|---|
| Saif-ur-Rehman et al. [42] | Tetraplegic patient | Utah Array | 192 | N/A | 30,000 | Human neural recordings for BCI applications | 98.90% Accuracy |
| Wang et al. [43] | Cynomolgus Monkey | NN-based Pattern Recognition | 128 | Varied | Brain areas | Classification across brain regions using neural networks | 100% Accuracy (cubic SVM) |
| Yi et al. [44] | Mouse | MEA, 64-channel electrode array | 49 | 27 | 30,000 | Multiple neural units from the 4mouse cortex | 93.1% Accuracy |
| Hall [15] | Rhesus monkey | Utah / Linear array | 96 (Utah) / 16 (Linear) | 2 | 30,000 | Comparative spike classification across brain regions | 86.23% Accuracy |
| Rácz et al. [45] | Wistar rats | Deep learning-based | 128 (Silicon probe) | 20+ | 20,000 | Evaluates deep sorting on high-class-count data | 89% Accuracy |
| Li et al. [46] | Macaque monkey | MEA | 128 | 3 | 24,400 | Standard benchmark for multi-class sorting in primates | 99.5% Accuracy |
| Sakamoto et al. [47] | Mouse | Tetrode | 32 | Varied | 32,000 | Real spike data with varied classes | 98.54 Accuracy |

## 4. Proposed AI-based techniques used for spike classification

Previously, some researchers tried to classify spikes in signal data using pattern recognition techniques. However, these techniques are not considered AI technologies. For instance, Biffi et al. developed an algorithm for detecting amplitude-threshold spikes, which was tested on both simulated and real signals using statistical analysis. They also developed a hierarchical classifier, which was evaluated on both simulated and real signals. The algorithm uses shape clustering and automatic classification, extracting , sorting , and classifying waveforms using a hierarchical classifier. The results confirmed the effectiveness of the clustering procedure [48].

Kaneko et al. proposed a method for detecting multi-neuronal spikes that combines multisite electrode recording, full pattern investigation, and hierarchical clustering. The approach is resistant to low-frequency sampling and noise because it encodes spikes as spatial damping vectors based on the covariance of a template and the signal that was detected at each recording station. This approach is more exact than other methods in classifying neural spikes [49].

Franke et al. presented a combined spike detection and classification algorithm for analyzing neuronal cooperation. The algorithm uses clustering and linear filters to improve the signal-to-noise ratio and introduces a method called "Deconfusion" for source separation. In this study, the authors achieved an Area Under the Curve (AUC) score of 80.88% [50].

Further AI-based research papers are presented below in the preprocessing methods, machine learning classifiers, and deep learning classifiers categories.

### *4.1 Preprocessing Methods*

The primary question in this section is: What effects does preprocessing have on successful classification?

Since this study reviews spiky data, data preprocessing can have a significant impact on the final classification outcome. For example, in a classification process using machine learning or deep learning, a preprocessing approach can serve as a recognizer for the classifier. In other words, classification requires first recognizing the incoming signal data, and a preprocessing method is the recognizer in this process. The essential input values for the signals can include time-domain and frequency features, as well as superposition-based features [32].

On the other hand, data preprocessing can be quite significant because it removes artifacts during the signal classification process. Artifacts, such as noise induced by electrode movement, pollute the source signal and interfere with the classification process [39].

No specific pipeline for signal preprocessing has been proposed yet [29]. However, the studies conducted in this paper demonstrate how preprocessing strategies might lead signal data to successful classification. Therefore, the available preprocessing techniques are presented below.

#### *4.1.1 Feature Extraction*

*A. Filtering*

One of the common methods for preprocessing is the use of filters. Filtering is usually applied as an initial and fundamental preprocessing step in neural spike classification pipelines. In many studies, low-pass, band-pass, and high-pass filters are utilized to eliminate Local Field Potentials (LFP) and high-frequency artifacts and improve Signal-to-Noise Ratio (SNR).

Ding et al., for example, employ a band-pass filter with a range of 10Hz to 3.4KHz to focus on specific frequencies, enhance SNR, and boost the distinctness and clarity of the spike features, improving spike classification accuracy [51]. Zacharelos et al used a low-pass filter to prevent aliasing by removing high-frequency noise and a high-pass filter to reject LFPs [52]. Moreover, Mohammadi et al. first apply a high-pass Butterworth filter to eliminate low-frequency components and then employ a local common average referencing filter to remove correlated noise shared among adjacent channels [53].

However, it is worth mentioning that the frequency bands should be chosen carefully. Otherwise, the shape and timing of spike waveforms can be negatively affected, and some useful information can be removed. Moreover, filtering alone cannot remove the noise that overlaps with spikes in the frequency band.

*B. Time Frequency Transforms*

The Short-Time Fourier Transform (STFT), Discrete Wavelet Transform (DWT), and Wavelet Transform (WT) and time-frequency-based techniques have been used for feature extraction [54]. These methods transform neural signals measured in the time domain into time-frequency representations, leading to both temporal and spectral feature extractions, which are essential for spike classification. These techniques help accurately classify signals by extracting relevant information. DWT is used to statistically evaluate the frequency bands of extracellular recordings, disclosing signal characteristics in both temporal and frequency domains [55]. Its capability for analyzing signals with different resolutions enables it to be highly effective for non-stationary signals such as neural spikes. The outcome also demonstrates that most algorithms have performed better when dealing with feature sets produced by wavelets [2]. Furthermore, the Fast Fourier Transform (FFT) focuses primarily on filtering techniques and divides the time domain into four frequency bands called alpha, beta, gamma, and delta. It can also transform time-domain signals into frequency-domain signals [56]. Despite its computational efficiency and usefulness for identifying and detecting main spectral features, its lack of time localization and assumption that a signal is stationary make it less effective for non-stationary neural spikes [57].

The DWT computes the WT fast by using sub-band coding. DWT is simple to install and reduces computation time and resource needs. Digital filtering techniques generate a digital time-scale representation of the signal. The signal is examined with filtering of varying cutoff frequencies and scales. The DWT signals are first put through an analysis filter bank and then decimated. These filters divide signals into two bands. The low-pass filter, like an averaging operation, extracts coarse information from the signal. The high-pass filter has a distinguishing action [58]. A band-pass filter can be employed to normalize data and eliminate noise [16]. This method facilitates the extraction of detailed frequency features, which can boost the classifier's ability to distinguish between different spike waveforms. However, the preprocessing method's effectiveness highly relies on the accurate selection of frequency bands and might need tuning for each dataset [59].

Haggag et al. proposed using the cepstrum to detect spikes in noisy signals. The cepstrum, calculated using the Inverse Fourier Transform (IFT), contains details concerning the intensity and amplitude of the original spectrum, separating meaningful features from noisy signals. It improves periodicity-related features and the robustness of the classifier in an environment with high noise, leading to enhanced performance and efficiency and reducing non-clustered spikes, with simulation results showing higher accuracy rates [60].

Yang et al. used both amplitude and phase components of the Fourier Transform (FT), allowing for fine differences within each cluster. They reported an accuracy of 99%, 95%, and 83% for three different noise levels [5]. This dual-component approach enhances the performance of spike discrimination, specifically across varying noise conditions. However, it might lead to computational complexity.

Besides, Haar DWT is used to extract features from the spikes or the noisy signals recorded from neuron sources in [37]. Due to its speed and simplicity, the Haar DWT works well and is effective for real-time use; however, it has limited capability for detecting detailed spike waveforms. In [61] STFT with one-dimensional convolution is applied to generate the spectrogram of the spikes and then a Meier filter is used to convert the spectrogram into a logarithmic Meier spectrogram to improve spike feature extraction and data reduction.

Sun et al. apply Scattering Convolution Networks to extract the time-frequency spike features from the recorded neural signals, using three key operations: WT, modulus operations, and low-pass filtering. WT is utilized to decompose the neural signals into multiple frequency bands. This enables them to capture spike features at different frequencies and time scales, which is essential for differentiating spikes with different durations and shapes. Next, the modulus operations approach is applied to the coefficient outputs. This non-linear operation preserves crucial amplitude information of the signal, while ensuring consistency despite small shifts in spike timing. This is essential because the occurrence of the spikes can vary slightly over time. Finally, a low-pass filter is used to reduce the sensitivity to noise and improve the SNR of spike signals. All of these approaches help improve robustness and spike classification performance [62].

For feature extraction, Dai et al. apply a window-gradient feature, a novel sample optimization approach. By emphasizing local waveform dynamics, this method improves spike classification performance. The proposed feature extraction method divides the spike signals into consecutive windows, calculates the gradient variations within them, and uses all the resulting features in place of the original waveforms. In other words, instead of the original waveforms, rapid changes in spike shapes are captured and utilized. This method allows them to highlight subtle differences in spike shapes and minimize the impact of noise and spike overlap. The results illustrate that this method helps enhance class classification, especially for signals with high noise levels [63]. However, the window width should be chosen carefully as a too large window may blur critical features, while a too small window may amplify sensitivity to noise. It needs to be tuned carefully.

#### *4.1.2 Dimensionality Reduction*

Adamos et al. use isometric feature mapping (ISOMAP) to reduce the dimensionality of the data while preserving the intrinsic geometry of the spike data. This preprocessing method is a nonlinear method for dimensionality reduction. Unlike PCA, which uses Euclidean distances, ISOMAP relies on the geodesic distances between data points to retain the manifold's true structure. This approach allows the authors not only to simplify spike classification even when spike overlaps exist but also to improve the robustness against noise and diversity in spike morphology, leading to an enhancement of spike classification performance [64]. Notably, ISOMAP is effective in conditions where spike waveforms include nonlinear variations, diverse noise levels, and spike overlaps. However, it heavily relies on an optimal selection of dimensionality and the neighborhood graph formation.

In [65]a new Discrete Derivative (DD) with two extrema sampling (DD-2Ex) algorithm is used for feature extraction and dimensionality reduction, representing spike signals using only four features. This method represents each spike using four features by computing the first-order derivatives of the spike at two different scales and then capturing the minimum and maximum values related to each scale. DD-2Ex extracts the essential geometric properties related to spike waveforms and provides high robustness to diverse noise levels while compressing data by a 16-fold reduction. This helps DD-2Ex be suitable for low-power, real-time spike classification.

Another method for data reduction is the use of Zero-Crossing Features (ZCF). This approach is based on the energy distribution of the spike waveforms. This method uses only two features to represent each spike waveform. It captures the spike waveform's energy on both sides (before and after) of its initial zero crossing. Yang et al. perform this method to reduce the dimensionality of spike data by over 90% while maintaining the nonlinear structure of the data [66]. These benefits make ZCF well-suited for energy-efficient and real-time neural recording platforms, though for complex spike shapes, it may be less effective.

Although the methods mentioned can be helpful in most datasets, recent research papers have indicated that with huge datasets, Principal Component Analysis (PCA) has a slight advantage [2]. PCA is considered both a feature extraction technique and a dimensionality reduction method. PCA is the most widely used dimensionality reduction approach. It is a robust transform technique for linearly transforming initial data into feature data with a reduced dimension while decreasing variance. Furthermore, it can cut processing expenses by converting a higher-dimensional feature space into a lower-dimensional principal component space. PCA identifies patterns in data and exposes similarities and distinctions, making it valuable in the recognition of faces, image reduction, and biological signal processing [67]. PCA simplifies and enhances spike classification by mapping high-dimensional spike data onto a small set of principal components, making it well-suited for large datasets. However, since it computes the linear distance between data points, it might lag behind other methods in nonlinear and high-noise conditions.

Wang et al. introduced a functional localization method in the brain of a cynomolgus monkey to identify spike patterns at different locations. The functional localization method is a multi-step pipeline containing spike recordings, preprocessing, dimensionality reduction, clustering (K-means), and spike classification. For dimensionality reduction, the authors apply PCA to transform the high-dimensional spike waveforms into reduced-dimensional data while retaining components representing 95% of the original variance. This method allows them to simplify the complex neural signals [43].

In [41], PCA is also applied to reduce spike dimensionality while retaining 85% of the data's variability.

In the study conducted by Okreghe, PCA is used to reduce high-dimensional spike data into a lower-dimensional one. It preserves 85% of the data's variability while reducing the data to 7-8 components [39].

In [68], Yang et al use a decision tree for spike classification. To feature extraction and reduction, they employ PCA and the newly discovered first and second derivative extrema (FDSE) algorithm and compare the results. The FDSE technique, in particular, performs denoising by concentrating on the critical regions in the spike waveform where rapid changes happen. These extracted minimum and maximum values of the first and second derivatives effectively capture the spike characteristics even in noisy conditions. Their results demonstrate that the proposed FDSE method outperforms PCA in high-noise conditions, and it is computationally simpler. However, it should be mentioned that in scenarios where overlapping spikes occur or spike waveform details are hardly distinguishable, the performance may be reduced.

#### *4.1.3 Clustering*

Numerous clustering algorithms have been reported in literature reviews, with a few being used in spike analysis algorithms. However, there is limited research on their suitability and compliance for spike analysis. Clustering methods are classified into seven categories: partitional, hierarchical, probabilistic, graph-theoretic, fuzzy logic, density-based, and learning-based. This categorization aids in comprehending the complexities of clustering algorithms. Clustering methods have evolved to better meet the subject under consideration and are adapted to be used in many fields [2].

In [65] k-Medoids algorithm is employed offline on a subset of the extracted features for clustering. It organizes similar feature vectors to put into distinct clusters corresponding to various spike classes. This approach enhances robustness against outliers by minimizing the distances between actual representative points (medoids) and data points, providing higher performance in noisy spike recordings compared to K-means. On the other hand, this method requires an estimation of the number of neurons, which might be unknown in real-world recordings. In addition, it must be run offline. Another limitation concerning K-Medoids would be its computational complexity. These limitations make this method unsuitable for real-time and on-chip implementations.

In addition, Adamos et al. apply Fuzzy C-Means (FCM) to cluster spike signals. It assigns a membership value to each spike signal, indicating its belonging to each group. Spikes that strongly belong to a particular group are chosen as reliable samples to train the classifier. Each spike, if exceeding a certain threshold, is allocated to the group with the highest membership [64]. Similarly, Ding et al. use the FCM for detecting spike clustering [51]. This method is useful for noisy recording data and is robust against spike waveform variation.

After preprocessing steps, such as filtering, denoising and normalization, K-means is another method employed for clustering. Park et al. [3], Okreghe et al. [39], Wang et al. [43], and Saif-ur-Rehman et al [41] apply this method to sort spikes into distinct clusters. They benefit from the clustering step to enhance robustness and performance, especially in the presence of different noise levels and waveform differences.

Moreover, in the study by Saif-ur-Rehman et al. [40], a combination of K-means and Cluster Accept Or Merge (CAOM) is utilized to sort detected spikes into distinct clusters. The CAOM method fine-tunes and improves K-means results by checking the consistency of the clusters and either accepting distinct clusters or combining similar ones.

#### *4.1.4 Feature Selection*

A larger number of extracted features can guide machine learning methods to represent information in a better and more widespread manner. Still, the problems associated with dimensionality may lead to inefficiency. To prevent this, a variety of techniques for lowering dimensionality and identifying variables have been created. Therefore, while designing computer-aided automated diagnosis systems, feature selection is crucial [55].

Noce et al. employed a multimodal method utilizing the Lilliefors-modified Kolmogorov-Smirnov test to choose the key features that best classify spikes. This optimization process aimed to automatically identify the best three features from a set of 12 features to facilitate better separation between spike types. The combination of the preprocessing steps, consisting of threshold-based spike detection, and the proposed optimized feature selection improved spike classification performance up to 95% [69].

### *4.2 Machine Learning-Based Classifiers*

#### *4.2.1 K Nearest Neighbors (KNN)*

KNN is one of the simplest and most straightforward yet practical classifiers for spike recordings, especially when integrated with feature extraction or clustering methods. KNN has been used in several studies using both real and simulated neural recordings to classify spikes, achieving high spike classification performance and often outperforming traditional approaches in accuracy and speed.

Sakamoto et al., used spectral clustering to group similar spikes into clusters using Laplacian eigenmaps, followed by a KNN to label the clusters, achieving high clustering and classification performance for both synthetic and real data from mice [47]. In another study, Hussein et al. applied PCA for feature extraction and supervised learning (support vector machines and KNN for the classification phase, demonstrating that KNN outperformed the other methods, achieving a classification accuracy of 100% on semi-simulated data at low noise and 94% at high noise levels [70]. Achieving 100% accuracy in this research is more likely due to the use of a noise-controlled, semi-synthetic dataset. It is potentially indicative of overfitting and may not generalize to the real-world signal recordings with more complexity and variability.

#### *4.2.2 Support Vector Machine (SVM)*

Due to their robust generalization ability, computational efficiency, and performance in linearly and non-linearly separable datasets, Support Vector Machines have been widely used among the classical machine learning methods to classify spike signals. According to a wide range of studies, SVM-based classifiers have achieved high accuracy rates, ranging from 90% to 100%, especially in low-noise and simulated spike signals.

Either independently or as part of hybrid classification approaches, SVMs have been utilized in various studies for spike classification. For basic SVM models, Kim et al. used nonlinear mapping in a spike train decoding algorithm via an SVM model, enhancing the performance of spike classification, despite the limitations of the use of simulated data and the limitations of the EEG-based brain–machine interface[71]. Ambard et al. utilized SVM with Postsynaptic Potential kernels and simulated spike patterns via a Leaky Integrate-and-Fire neuron model, achieving robust classification accuracy of spike patterns into target and background patterns [72].

As for integrating SVM models with preprocessing methods, Ding et al., for example, used a multi-class SVM approach with the “one-against-all” method, which is trained by data obtained from the output of the FCM algorithm, demonstrating high spike performance even under challenging scenarios such as various signal-to-noise ratios and superimposed waveforms [51]. Similarly, Noce et al. utilized a multimodality test for feature selection before a supervised SVM classifier, obtaining a high spike classification accuracy of about 95% to classify at most 10 units [69].

To boost spike classification performance under various noise levels, SVMs have been integrated with various hybrid models. For example, Radmanesh et al. combined a Deep Contractive Autoencoder (DCAE), followed by a clustering algorithm using K-means and a classification method using SVM, achieving over 90% accuracy under various noise levels [73]. Dai et al. proposed a sample optimization method called Window Gradient for feature extraction and an SVM for classification to handle challenges related to high noise levels and waveform similarities. The proposed method, despite the larger running time, outperformed PCA-based approaches and improved spike classification accuracy significantly, particularly when the sample similarities and noise levels are large [63].

In real-world spike recordings, Wang et al. implemented MEAs in some cynomolgus monkeys' brains and recorded the neural activities and applied cubic SVM for classifying spike patterns and identifying electrodes' locations in monkeys, reaching 100% accuracy of spike classification in normal monkeys and 94.5% accuracy in Parkinson's disease monkeys [43]. Yang et al. proposed a methodology starting with signal preprocessing, followed by a PCANet utilized for eigenvalue feature extraction from the spikes and subsequently using a SVM to classify neural spikes on high-volume multi-electrode recordings into their original units, achieving precise spike sorting with low computational complexity [74].

Moreover, a Gini-based SVM was implemented by Vogelstein et al., consisting of removing noise from the spikes in the detection step and a supervised SVM trained using simulated labeled data, achieving higher classification accuracy over different noise levels [75].

In conclusion, the Support Vector Machines results show that they are successful in spike classifications. However, they might experience low performance in cases such as significant spike overlap or high channel recordings.

### *4.2.3 Boosting Methods*

Another method that has been implemented for spike classification was Boosting algorithms, particularly Extreme Gradient Boosting (XGBoost). Boosting algorithms, although less frequently applied, due to their capability of capturing complex patterns, could obtain strong classification accuracy, especially when combined with robust feature extraction methods.

For example, Sharifrazi et al. proposed a mix of convolutional neural networks and XGBoost, and PCA for the preprocessing part for spike classification for the detection of Zika and Dengue disease, achieving 95.3% accuracy, precision, recall, and F1 Score [13]. In addition, Eom et al. integrated an ensemble of Autoencoders (AE) to extract features with some unsupervised machine learning algorithms such as K-means and the Density-Based Spatial Clustering of Applications with Noise (DBSCAN), achieving high spike classification on both simulated and in vivo datasets under various noise levels [76].

#### 4.2.4 Other ML Methods

Several other machine learning methods- such as graph-based models, self-organizing networks, decision trees, and shallow neural architectures- have been used for spike classification. To improve performance, they have been combined with various preprocessing methods, such as feature extraction and dimensionality reduction techniques.

A graph network, particularly a GEMsort method, was introduced by Mohammadi et al. It clustered neural spikes by analyzing features such as channel positions and temporal profiles, achieving high sorting accuracy on both simulated and experimental datasets [53]. In a separate study, Yang et al. introduced an on-chip system for spike classification using a Self-Organizing Map (SOM) algorithm and Zero-Crossing Features for feature extraction, performing spike classification with high classification accuracy over various levels of noise interference [66].

For other lightweight classifiers, Yang et al. demonstrated a Decision Tree (DT)-based spike classification approach to reduce memory usage and system area and optimize computation complexity without compromising performance [68]. Meanwhile, Adamos et al. utilized a combination of ISOMAP and FCM for dimensionality reduction and clustering, and Extreme Learning Machine (ELM) for classifying unclassified spikes [64].

These aforementioned methods offer promising performances, especially in scenarios where conventional models are not successful in classifying spikes or are limited by hardware. However, they are not frequently implemented in high-density spike recordings.

### 4.3 Deep Learning-Based Classifiers

#### 4.3.1 Deep Neural Network (DNN) and Artificial Neural Network (ANN)

In recent years, several studies have used Deep Neural Networks (DNNs) and Artificial Neural Networks (ANNs) to classify spikes, achieving strong performances under various conditions.

For example, Chandra et al. employed a supervised ANN and trained the model using spike templates with labels, including both individual and overlapping spikes. Compared with the matched template filter method, the proposed method demonstrates high detection and classification accuracy, even in challenging scenarios such as spike superpositions [77]. Similarly, Zacharelos et al. used a supervised dense neural network trained on simulated extracellular recordings generated by NeuroCube. While requiring low energy and minimal silicon area, the proposed model achieved high accuracy and is suitable for on-chip implementation [52].

Mirfakhraei et al. presented a three-layer neural network method to classify action potentials (spikes) in multi-unit extracellular recordings from cats' peripheral nerves, successfully classifying approximately 46% of individual units — performing significantly better than the traditional template-matching method (16% classification accuracy) [78]. Meyer et al., In another study, employed a simple single-hidden-layer neural network to classify noisy spike waveforms and achieved a spike classification accuracy between 89% and 100% on manually generated data with different noise levels (ranging from 0 to 50 %), showing that complex network architectures are not always needed for achieving basic spike classification and simple networks are also able to perform this task effectively. However, accuracy may decrease in real data with overlapping spikes, and more advanced architectures are required for such complex data [79].

Another novel spike classification pipeline using deep learning, called SpikeDeep-Classifier, is described by Saif-ur-Rehman et al. They combined PCA for data dimensionality reduction and some Deep Learning models for clustering (K-means and CAOM), reaching 88.03% and 86.95% average accuracies on human and non-human datasets recorded via different types of electrodes and other recording systems, respectively [41]. In addition, Sun et al. enhanced spike classification by using a Scattering Convolution Network to extract features and a lightweight convolutional neural network to classify neural spikes, gaining a high accuracy of up to 99.97% and about 92% computation reduction on extracellular recordings with different levels of noise interference [62].

Wu et al. introduced a Few-shot spike sorting approach combining DNN, GAN, and Adversarial representation learning to classify action potentials with a small number of labeled spikes to handle the issue of limited labeled data, demonstrating high classification performance across varying signal-to-noise ratios on in vitro and synthetic extracellular recordings [80].

*4.3.2 Multi-Layer Perceptron (MLP)*

MLPs, thanks to their robustness in handling overlapping spikes, noise, and unlabeled data, have been widely used for spike classification.

Issar et al. built an ANN to identify spikes from noise in EEG recordings instantaneously. The method improved decoding performance for most sessions, even while data quality declined with time, highlighting the robustness of MLPs against signal quality degradation [81].

In contrast, Park et al. proposed a deep learning-driven spike-sorted technique to create pseudo-labels for training the MLP in cases where labeled data is unavailable, achieving high classification accuracy on difficult spikes. This method demonstrates high effectiveness in detecting spikes recorded from noisy single-channel extracellular signals [3].

Wouters et al. concentrated on addressing the problem of overlapping spikes in the feature space by proposing MLP architectures to generate a feature map, which effectively addresses spike overlaps in extracellular recordings. The results demonstrated high recall and precision in spike classification, showing the MLP's capability to manage challenging conditions, such as spike overlap, effectively [82, 83].

Lotfi et al applied a transformer network for feature extraction, followed by a supervised MLP for neural spike classification to improve spike classification. This hybrid method achieved an average of 99.85% classification accuracy on the simulated dataset and 95.06% on the real dataset, showing the impact of prepossessing and informative features on MLP performance for challenging scenarios like noise interference and overlapping spikes [84].

*4.3.3 Convolutional Neural Network (CNN)*

Thanks to their robustness in feature extraction, CNNs are another architecture that has been frequently utilized for spike classification, particularly in challenging conditions such as overlapping signals, noise signals, and real-time processing analysis.

Hall's study showed that replacing fully connected layers with CNNs for spike classification can improve spike classification accuracy by 89.5% across multiple cortical areas, showing CNNs' ability to identify electrode waveforms like spikes or noise [15].

Zhang et al. progressed this method further by proposing a combination of STFT to transform spike signals into spectrogram representations and a 4-layer CNN to classify overlapping and individual spikes, surpassing traditional template matching and hybrid CNN-LSTM models with an average F1 score of 96.13%, an average precision of 97.06%, and an average classification accuracy of 95.74% [61].

To deal with time-dependent features and enhance the robustness of spike classification, multiple studies have utilized a combination of CNNs with LSTM architectures. Wang et al., for instance, used a one-dimensional CNN to extract spike samples and filter out background activity samples, and a combination of LSTM and two-dimensional CNN performs as a spike classifier, resulting in classification accuracies of 99% and 95% for simulated and experimental datasets, respectively [85]. Similarly, Rácz et al. combined CNN and LSTM and achieved an average accuracy of 89% for spike classification generated by over 20 distinct neurons [45]. Liu et al. also focused on handling overlapping spikes, reporting 99.9% and 96.18% for individual spikes and 99% and 88.67% for overlapping spikes for simulated and real-time data, respectively [86]. The results show the robustness and effectiveness of hybrid architectures.

In contrast, Li et al. applied a supervised one-dimensional CNN approach for spike classification in extracellular neural recordings and still obtained average classification accuracies of 99% and 96.53% for simulated and real datasets, respectively, showing that using a pure but careful CNN architecture can be sufficient to achieve high performance in challenging situations with overlapping spikes and high noise levels [46].

To enhance overall spike classification performance and robustness, several other researchers have integrated CNNs with multi-stage pipelines. For example, Okreghe et al. presented Deep Spike Detection (DSD) using CNNs for channel selection to identify neural event recording channels and filter out artifact channels, PCA for feature extraction, and K-means to classify spikes, achieving the overall classification accuracy of 91.53% on the simulated data [39]. In addition, Saif-ur-Rehman et al. introduced an adaptive SpikeDeep-Classifier with a combination of CNN and K-means + CAOM, obtaining spike classification accuracy ranging from 87.7 to 92.74% on real and simulated datasets [40].

CNNs have also demonstrated efficient performance in hardware implementations. As an example, Yi et al. proposed a lightweight CNN trained by generated datasets with labels via MountainSort. The quantized model reduced storage requirement by 93% while achieving an impressive spike classification accuracy of 86.1% [44].

Overall, based on the results of the studies above, CNNs are generally accurate; however, when combined with preprocessing techniques, adaptive designs, or time-aware models, they perform better, especially in challenging scenarios like noisy, overlapping, or multi-source environments.

#### *4.3.4 Spiking Neural Network (SNN)*

Spiking Neural Networks (SNNs), due to their suitability for recognizing temporal patterns and energy efficiency, are another subset of artificial neural networks that have been increasingly used for spike classification in several studies.

Mukhopadhyay et al. combined an unsupervised K-means algorithm with a supervised SNN approach to boost spike classification accuracy. The hybrid model was also designed to retrain itself when the distance between the new neural spike and its nearest cluster exceeded a predefined threshold, achieving high classification accuracy while consuming low power [87].

Cheok proposed an SNN-based spike classifier trained by the unsupervised Spike-Timing-Dependent Plasticity (STDP) rule. To map each cluster to its specific spike class, a post-training phase was applied using labeled data, achieving an average spike classification accuracy of 93.31% on the simulated dataset and showing high robustness under different levels of noise [88].

Following this, Bernert et al. developed a spike sorting and classification algorithm using attention mechanisms and STDP into their SNN architecture. The network managed the rare occurrences of spikes and Intrinsic Plasticity to adapt neuron thresholds to handle patterns of varying lengths, achieving high F1-score and recall, especially at low signal-to-noise ratio on both simulated and real datasets [89].

*4.3.5 Binarized Neural Networks (BNN)*

To reduce memory requirements and computational complexity while maintaining acceptable accuracy, several studies have utilized Binarized Neural Networks (BNNs) for spike classification, particularly in real-time and hardware-constrained environments.

To achieve significant power and area efficiency, Valencia et al. implemented BNNs with binarized weights and activation functions across a combination of real and synthetic neural signal datasets with various noise levels. The average classification accuracy of the BNN-based classifier was 95.25% for the Easy1 dataset and 82% for the difficult2 dataset, showing the need for a trade-off between efficiency and accuracy [38].

Valencia et al. also developed a hybrid approach that first applied unsupervised K-medoids clustering to generate initial labels, followed by supervised training of a partially Bayesian Neural Network (BNN). The method achieved high accuracy under challenging, noisy conditions on Wave_Clus datasets, showing the effectiveness of combining preprocessing techniques with lightweight models [65].

*4.3.6 Other DL Methods*

In addition to the DL methods mentioned earlier, several other DL strategies have been applied for spike classification, particularly in scenarios involving overlapping spikes, computational efficiency requirements, or class imbalance.

Reddy et al. utilized a combination of MLP and Deep Belief Networks (DBN). They implemented WT as a preprocessing method for spike classification to understand the neuronal activity of extracellular recordings. They achieved a spike classification accuracy of 92.56%, showing the effectiveness of combining traditional methods with deep generative methods such as DBNs [37].

Since the data recorded via MEAs suffers from multiclass disequilibrium (spike overlapping and variations in neuronal firing rates), Li et al proposed a framework combining deep Reinforcement Learning (RL) and Dynamic Reward Function (DRF) to handle these issues in spike classification. Spike sorting was modeled as a Markov sequence decision, with the DRF designed to enhance the focus of the agent on underrepresented classes. To train and test the model, both the Wave_Clus and macaque datasets were used. The proposed model achieved an accuracy of 98.4% and a F1-score of 92.8% on the Wave_Clus dataset and an accuracy of 97.9% and a F1-score of 95.8% on the macaque dataset, demonstrating robustness against different noise levels and overlapping [90].

In another study, Meyer et al. proposed a novel end-to-end method called Dual Sort, which utilizes a running neural network —a simple Neural network combined with downstream post-processing —to detect and classify neural spikes, obtaining high performance as a spike classifier, with an average classification accuracy of 98.04% and a Macro F1 score of 98.53% on the synthetic data and a classification accuracy of 95.22% and a Macro F1 score of 96.74% on the experimental data while maintaining simplicity [91].

So far, in each subsection, a brief comparison of relevant studies has been provided. In the descussion section, a more in-depth and detailed comparison will be conducted based on distribution, data type, and performance.

### *4.4 Evaluation Metrics*

Using commonly used assessment measures for the classification tasks, the effectiveness of the proposed methods for spike classification is assessed. The measurements include accuracy, precision, recall, sensitivity, specificity, F1-score, and AUC, each providing a unique perspective on the model's performance. Based on these metrics, we are wondering which machine learning technique achieves the best overall performance. The equations of these metrics are available in [48, 92, 93].

### *4.5 Data mapping of included studies*

In this section, the research papers selected by various databases are analyzed in terms of their trends over time and publisher contributions in this realm.

According to Figure 5, the graph is characterized by an increasing trend between 1994 and 2024. The importance of spike classification using AI-based methods was not discovered before 2017. As a consequence, a steady progress of publishing research papers has been observed from 1994 to 2017. Therefore, the timeline was truncated to reflect modern AI relevance. Between 2017 and 2022, a slight fluctuation in this area was seen, and the spike classification issue increased dramatically from 2022 to 2023. That is, this topic reached a peak in 2023, with 11 research papers published in just one year. Spike classification experienced a slight drop, followed by growth up to the present.

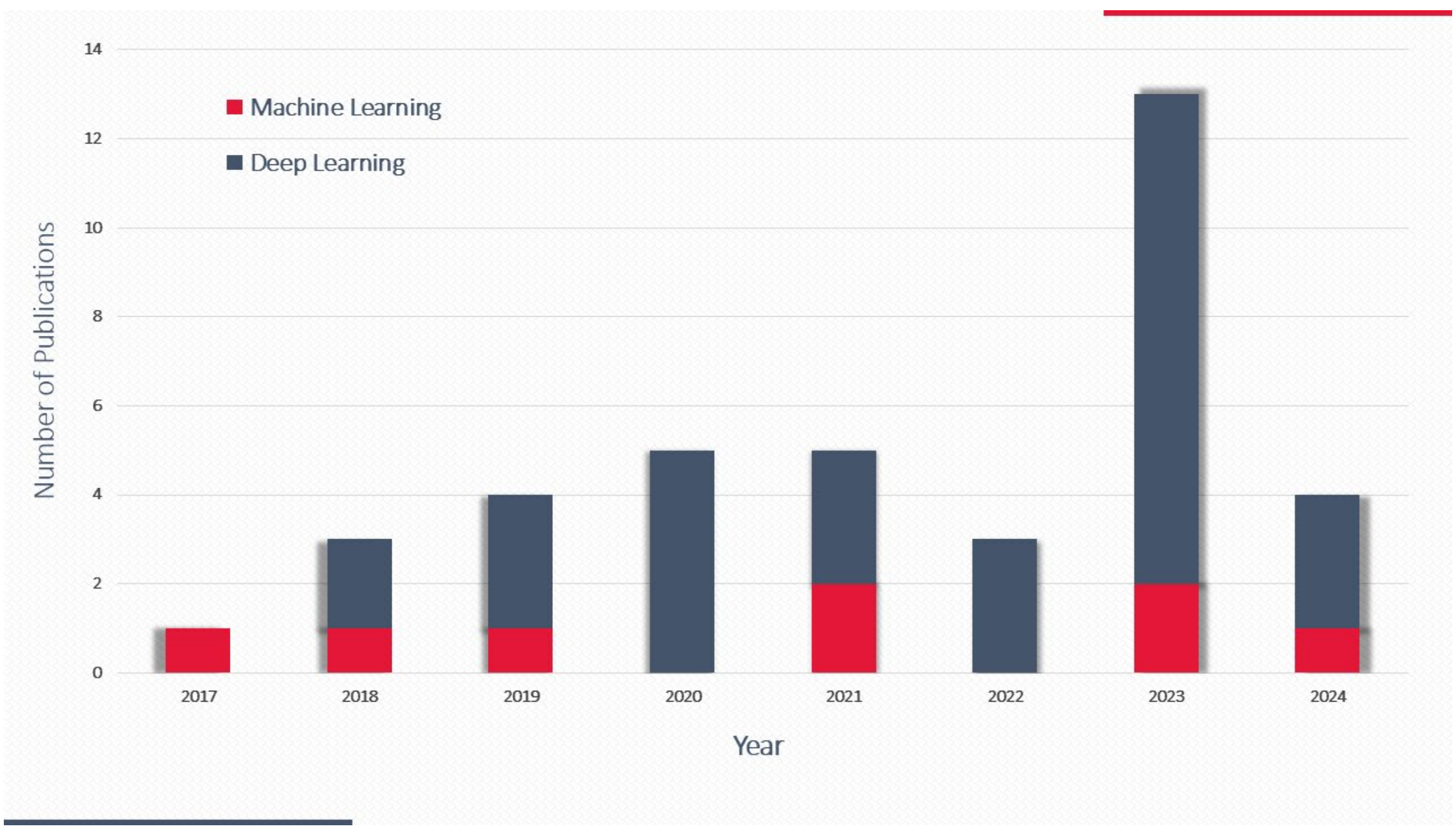

Figure 5. Bar chart shows the number of papers published between 2017 and 2024. . Before 2017, fewer AI-based studies were conducted.

# 5. Discussion

## *5.1 Distribution analysis*

In this section, we show approaches followed by the papers published in the field of spike classification. The current research papers are compared. It is also evident that, based on the AI-driven preprocessing and classification methods introduced in Section 3, certain gaps remain in the spike classification task. What are these remaining challenges? This case is discussed in detail in the conclusion section. Figure 6 depicts the frequency of use for each method based on the research papers selected for this topic.

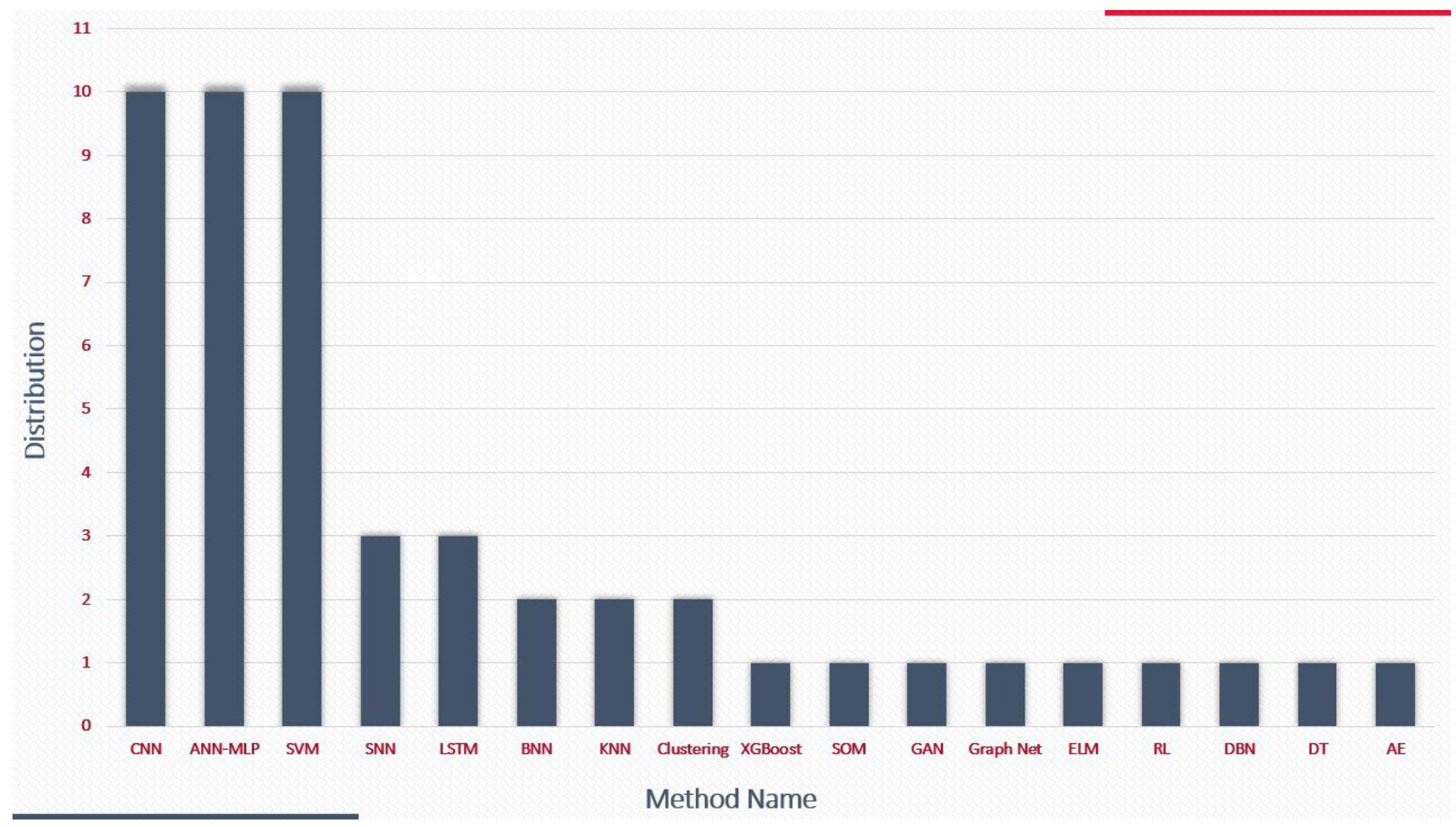

Figure 6. Bar chart shows the distribution (number) of AI methods used for spike classification.

****CNN**: Convolutional Neural Network, **ANN-MLP**: Artificial Neural Network - Multi-Layer Perceptron, **SVM**: Support Vector Machine, **SNN**: Spiking Neural Network, **LSTM**: Long Short-Term Memory, **BNN**: Bayesian Neural Network, **KNN**: K-Nearest Neighbors, **XGBoost**: Extreme Gradient Boosting, **SOM**: Self-Organizing Map, **GAN**: Generative Adversarial Network, **Graph Net**: Graph Neural Network, **ELM**: Extreme Learning Machine, **RL**: Reinforcement Learning, **DBN**: Deep Belief Network, **DT**: Decision Tree, **AE**: Autoencoder.

At first glance, it appears that deep learning methods are more popular than machine learning approaches.

Figure 6 shows that deep learning methods, CNN and ANN/MLP, and SVM hold the first three spots in the distribution number.

In the diagram, it can be observed that CNN, MLP/ANN, LSTM, and SNNdeep learning methods, occupy the initial positions of the distribution. Moreover, GAN, RL, and DBN have gotten less attention than other deep learning algorithms. Since deep learning methods are generally popular, it seems that LSTM has the potential to attract more attention from researchers in the future. These strategies are easily integrated with more widely used procedures like CNN and MLP moreover, attention-based mechanisms can be one of the prominent algorithms that can be utilized in the future.

Figure 6 indicates that SVM is commonly employed in machine learning methods. Further approaches, such as SOM and DT, are also less prevalent.

### *5.2 Preferred techniques for each data type*

The review of the datasets presented in Section 2 and the classification techniques in Section 3 can be a promising starting point for an in-depth analysis of spike classification. Given the dataset information presented in Table 1, it seems that the type and recording technique can be suitable indicators for dividing the datasets into different categories. Accordingly, in this section, we divide the datasets into categories “High-Channel MEA, Tetrode and Utah”, “Low-Channel MEA, Tetrode and Utah”, and "Simulated Datasets Excluding MEA, Tetrode and Utah". Then, we evaluate the classification techniques on each category and state which classification technique is more suitable for each type of data.

In papers that have used high-channel datasets (Category 1), due to the high complexity of the data and the difficulty in recognizing the shape of the data in such datasets, more complex classification techniques have been mainly used. In most of the publications, models based on deep learning have been implemented. This type of dataset, the primary focus has been on deep learning-based techniques, namely ANN, CNN, SNN, BNN, RL, GAN and attention-based mechanisms, for spike classification. However, simpler machine learning models such as KNN and SVM have also been considered to a lesser extent.

In the papers that use low-channel datasets (Category 2), simpler machine learning models have been used effectively due to the lower complexity of the datasets and the limited size of the samples. In this regard, simpler machine learning techniques such as PCA, SVM, ELM, and Clustering can be mentioned.

For simulated datasets (Category 3), it seems that the diversity of this type of data is very high, depending on the institution where this data was created. Therefore, both machine learning models and deep learning hybrid models have been widely used. In studies where the data was less complex, several papers introduced SVM as the main classifier. Also, in publications where the data was more complicated, deep learning models, including ANN, SNN, and a hybrid model called CNN-LSTM, were given more attention.

Based on the explanations provided in Table 3, the different categories of datasets and the techniques utilized for each group can be seen separately in their publications.

Table 3: Dataset categories discrimination based on their types and preferred classification techniques for each.

| *Category* | *Preferred Techniques* | *Example Studies* | *Descriptions* |
|---|---|---|---|
| 1. High-Channel MEA, Tetrode and Utah | Deep Learning Methods including ANN, CNN, SNN, BNN, Deep RL, GAN and Attention Mechanism | [13], [15], [81], [41], [46], [90], [44], [87], [65], [89], [83], [42], [80] | Large-Scale Recordings and Complex Datasets |
| 2. Low-Channel MEA, Tetrode and Utah | Machine Learning Methods Including Clustering, PCA, SVM and ELM | [64], [73], [76], [75], [74] | Limited Sample Sizes, Learning from Features |
| 3. Simulated Datasets Excluding MEA, Tetrode and Utah | Machine Learning, especially SVM, Deep Learning, especially CNN-LSTM | [71], [61], [85], [86], [72], [88], [79], [69], [70], [63] | Large-Scale and Small-Scale Datasets- based on Their Source |

### *5.3 Performance analysis*

As can be observed, the application of machine learning and deep learning algorithms for spike classification is expanding. So far, it has been identified what results each machine learning and deep learning method has yielded. In this part, we will look at how these strategies function.

Among the presented machine learning methods, KNN, SVM, and XGBoost have the best performance. All these strategies achieved an accuracy of 95% in the worst scenario and 98% in the best situation, approximately. Figure 7 depicts which strategy performed better in terms of accuracy. It should be noted that most of these methods have been employed in several studies, and Figure 7 displays the highest accuracy derived from these papers. The KNN method has the highest reported accuracy among all machine learning approaches in comparison to other machine learning methods. The combination of clustering with other machine learning methods has demonstrated satisfactory results and has been widely used in several studies in recent years.

The deep learning methods SNN, CNN, and CNN+LSTM have yielded the best results. All the presented accuracies in these methods are greater than 96%, with the best cases being above 99%. Deep learning approaches have also lately been employed in various studies, yielding excellent results. Figure 7 demonstrates that deep learning approaches have generally outperformed machine learning methods. Furthermore, Figure 6 indicates that the use of deep learning approaches is on the rise.

It should be emphasized that although all these strategies used spiky signal data, the datasets used in these papers varied. It is also worth noting the importance of preprocessing in this context. Each study implemented a distinct data preprocessing approach based on the available dataset. As indicated in the introduction, preprocessing procedures serve as recognizers for classifiers. As a result, the type of preprocessing utilized in each study has a dramatic impact on the final classification using machine learning and deep learning techniques.

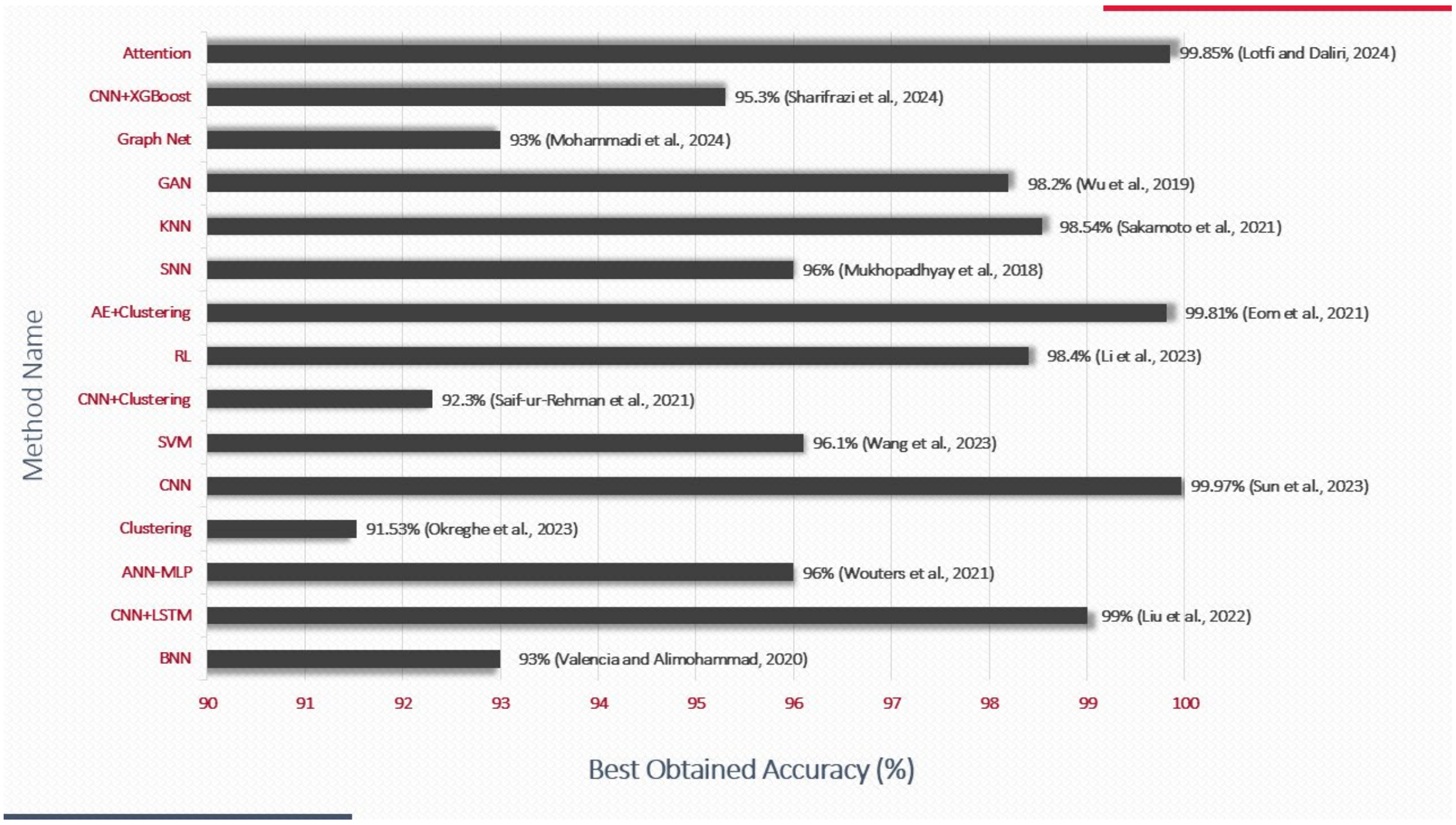

Figure 7. Distribution of utilized machine learning and deep learning methods related to selected research papers for spike classification.

****CNN+XGBoost**: Convolutional Neural Network + Extreme Gradient Boosting, **Graph Net:** Graph Neural Network, GAN: Generative Adversarial Network, **KNN:** k-Nearest Neighbors, **SNN:** Spiking Neural Network, **AE+Clustering:** AutoEncoder + Clustering, **RL:** Reinforcement Learning, **CNN+Clustering:** Convolutional Neural Network + Clustering, **SVM:** Support Vector Machine, **ANN-MLP:** Artificial Neural Network - Multi-Layer Perceptron, **CNN+LSTM:** Convolutional Neural Network + Long Short-Term Memory, **BNN:** Bayesian Neural Network.

Based on the results presented and by examining the performance of different methods, it is clear that deep learning models outperform machine learning models. In most papers, it has been indicated that newer deep learning methods have demonstrated better performance than traditional machine learning models and older deep learning models. For instance, the combination of MLP with the Attention Mechanism has been able to achieve the highest accuracy (99.85%), among all machine learning and deep learning methods [84]. Obviously, in the scope of spike classification, the goal is to train the spikes in the signal data, and the Attention Mechanism, by having the ability to pay attention to a specific part of the data, has been able to handle this task well.

Moreover, other deep learning models, such as CNN and its combinations, such as CNN-LSTM or CNN-XGBoost, have so far achieved very acceptable outcomes compared to other machine learning models. So that in all cases, they have reached an accuracy of over 95% [13, 61, 85, 86].

Further deep learning methods have also been able to provide relatively suitable results. For example, although methods AE and DBN did not achieve as high accuracy as newer deep learning models, such as the combinations of CNN and Attention-based models, they can still be suitable options with the automatic feature learning mechanism for complex and large datasets, like all deep learning models. Additionally, other deep learning models that have received less attention, such as GANs [80] and RL [90], can have their special applications in spike classification, which are suitable options for huge, complex datasets.

In contrast, traditional machine learning models such as SVM and KNN demonstrated poorer performance than deep learning models. These models can be suitable options when the desired dataset is well-structured and benefits from limited samples, due to their simplicity and higher learning speed than deep learning models. However, in situations where the dataset is complex and large (which is often the case nowadays), they perform worse than deep learning models.

In Section 6, the challenges in spike classification, how to provide solutions, and future perspectives are discussed in detail, based on the studies reviewed so far.

## 6. Open challenges and future insights for spike classification

### *6.1 Key challenges*

#### *6.1.1 Computational cost*

One of the most significant issues we encounter in the realm of spike classification is the computational cost. Many low-power gadgets do not support pricey approaches. Currently, deep learning approaches have a large computational cost [19]. Some academics recommend using several Linux-based PCs to improve this difficulty, while others find SNNs useful. Although some claim that using the SNN method in spike classification can cut calculation costs, other researchers think that this issue is still a challenge in this field and the SNN approaches, like other deep learning methods, are incapable of reducing effective energy [94].

#### *6.1.2 Real-time classification*

Another basic challenge in spike classification is the quick and real-time processing of the data [67]. This is an essential issue for hardware implementation, online processing, and data storage capacity reduction [48]. Some researchers employ various strategies to lower system computation. However, some scientists feel that extracting effective information features is complex, especially for multi-channel data. To overcome system complexity challenges, obtain a real-time classification system, and improve the speed of the classification process, some have argued that effective preprocessing can be severely fruitful. Of course, the preprocessing is linked to complexity reduction. For instance, dimensionality reduction, feature selection, and feature extraction methods like the discrete wavelet transform [95, 96].

#### *6.1.3 Odd spike shapes*

Another significant topic to consider is robustness in signal interference and accurate spike classification [5]. Some signals are challenging to classify because of their complicated patterns [97]. High-throughput classification of trustworthy biomarkers is especially crucial [98]. The overlapping of spikes makes classification difficult [49, 50]. Some spikes have unusual forms that make them difficult to classify. The data's complex structure may lead the final classifier to malfunction. For example, spikes with sharp waves are more difficult to classify than usual shapes. It is also exceedingly difficult to recognize frequent spikes in signal data [99].

*6.1.4 Sufficient and reliable data*

One of the primary issues we encounter in the field of classification of spikes is the question of the number of valid data points, especially in the case of patient data [93]. Therefore, some of the datasets are imbalanced. Many of the proposed methods are more suited for the classification of balanced datasets, and their implementation on imbalanced datasets may be ineffective [21]. Insufficient data remains a challenge for machine learning techniques in spike classification to operate properly [100], especially for deep learning methods. They require more data than machine learning methods, and the quantity and quality of this data can drastically impact the performance of classification [3]. Another issue with accurate data is that many studies and proposed methodologies have been undertaken using data from a single institution. Some approaches may not perform well on different datasets belonging to other institutions [16]. As a result, it is impossible to provide a definitive result about their performance on other datasets, and the situation may remain unclear.

### *6.2 Future directions*

The aforementioned challenges influence the future directions. In other words, based on the present studies, it appears that most scientists are attempting to address the current issues in future work to some extent.

One of the future directions is the employment of hybrid methods to improve the efficiency of present methods, particularly for outlier data [21]. Specifically, possible future directions may include the integration of transformer-based attention modules with traditional convolutional or recurrent layers for better spatial and temporal pattern recognition from complex biological signals. The performance of such deep learning methods was compared in the discussion section.

It is worth noting that using hybrid methods may increase computational cost. The existing limits can be solved by increasing the hardware capacity [101]. Some academics claim that parallel processing can reduce computational costs and enable a real-time technique [100]. Consequently, other possible studies may focus on minimizing the computational costs. For instance, utilizing pre-trained models in future studies can be fruitful. Although pre-trained models, especially large models, lead to expensive initial training costs and hardware requirements for their producer companies, using these models by fine-tuning on private datasets can be significantly efficient for many researchers by eliminating the need to train models from scratch [102].

As previously indicated, many current datasets lack the efficiency required for effective classification. Future work may include dataset improvement and classification. For example, some scientists hope to improve the classification process by increasing the number of labels [103] or the number of features [96]. Other researchers believe that increasing the number of real samples of participants will result in more accurate classifications in the future [93, 104]. Further studies claim that employing data augmentation methods to increase the number of samples and balance datasets will result in improved accuracy [105]. Thus, constructing large-scale datasets to improve the generalization and robustness of future studies can be critical. In this way, utilizing AI-based procedures such as generating synthetic data samples as an augmentation technique and semi-supervised learning for using a large number of unlabeled samples to enhance the model's performance can be influential.

Some researchers claimed that to improve the accuracy of the presented approaches, future work could focus more on data preprocessing to eliminate signal artifacts [102] or detect outliers [106]. But another major component in advancing this area is the development of standardization protocols. There is great variation in the ways data are collected, preprocessed, labeled, and evaluated across different studies; hence, results cannot be compared effectively. An appropriate direction for future research would be to involve shared benchmarks and unified preprocessing pipelines, such as consistent filtering plus artifact removal techniques. This would make results much more comparable between different studies and hence, would advance collaboration as well as accelerate progress.

## 7. Conclusion

As mentioned, AI and neuroscience are intimately intertwined. Previously, scientists analyzed signal data without employing AI technologies. However, this work illustrated that AI can significantly impact the processing of this type of data, providing researchers with more accurate results. We addressed the issue of spike classification in signal data and demonstrated the effectiveness of using AI-based preprocessing and classification methods in the results.

In spike classification, it was mentioned that specific spike characteristics are employed for classification. Other researchers who apply AI approaches divide the procedure into three parts: preprocessing, classification, and evaluation. In the preprocessing section, the effective methods were thoroughly outlined, accompanied by relevant examples. In the evaluation part, the important metrics were also listed. Then, the findings of each study using these metrics were expressed. In the classification section, AI algorithms were classified into two main types: machine learning and deep learning-based approaches.

Overall, this review introduces the reader to the notion of spikes and how they appear in signal data. It introduced the available datasets and the importance of AI methods in the preprocessing and classification of spikes. Eventually, the research findings were assessed, and key challenges and future directions were identified.

## Funding

This publication received no external funding.

## Conflict of Interest

The authors declare that there is no conflict of interest regarding the publication of this paper.

***Appendix 1***. *Abbreviations used in this review.*

| *Abbreviation* | *Complete Form* | *Abbreviation* | *Complete Form* |
|---|---|---|---|
| **AI** | Artificial Intelligence | KNN | K Nearest Neighbors |
| **EEG** | Electroencephalography | MEA | Multi-Electrode Array |
| **BCI** | Brain-Computer Interface | SVM | Support Vector Machine |
| **MEG** | Magnetoencephalography | DCAE | Deep Contractive Autoencoder |
| **PRISMA** | Preferred Reporting Items for Systematic Reviews and Meta-Analyses | XGBOOST | Extreme Gradient Boosting |
| **IEEE** | Institute of Electrical and Electronics Engineers | AE | Autoencoder |
| **IOP** | Institute of Physics publishing | DBSCAN | Density-Based Spatial Clustering of Applications with Noise |
| **ACM** | Association for Computing Machinery | DT | Decision Tree |
| **MDPI** | Multidisciplinary Digital Publishing Institute | SOM | Self-Organizing Map |
| **AUC** | Area Under the Curve | ELM | Extreme Learning Machine |
| **LFP** | Local Field Potentials | ANN | Artificial Neural Network |
| **STFT** | Short-Time Fourier Transform | DNN | Deep Neural Network |
| **DWT** | Discrete Wavelet Transform | GAN | Generative Adversarial Network |
| **WT** | Wavelet Transform | MLP | Multi-Layer Perceptron |
| **FFT** | Fast Fourier Transform | CNN | Convolutional Neural Network |
| **IFT** | Inverse Fourier Transform | LSTM | Long Short-Term Memory |
| **FT** | Fourier Transform | DSD | Deep Spike Detection |
| **ISOMAP** | Isometric Feature Mapping | SNN | Spiking Neural Network |
| **ZCF** | Zero-Crossing Features | STDP | Spike-Timing-Dependent Plasticity |
| **PCA** | Principal Component Analysis | BNN | Binarized Neural Network |
| **FCM** | Fuzzy C-Means | DBN | Deep Belief Network |
| **CAOM** | Cluster Accept or Merge | RL | Reinforcement Learning |

***Appendix 2***. *Summary of selected relevant research papers in this review.*

| *Study* | *Preprocessing Method* | *Classification Method* | *Performance (%)* |
|---|---|---|---|
| Lotfi and Daliri, 2024 [84] | Band-pass filter + Feature Extraction | MLP attention-based mechanism | 99.85 |
| Sharifrazi et al., 2024 [13] | PCA | CNN + XGBOOST | 95.3 |
| Mohammadi et al., 2024 [53] | High-pass filter + PCA | Graph Net Clustering | 93 |
| Hussein and Mohammed, 2024 [70] | Band-pass filter + PCA | SVM / KNN | 90-95 / 94-100 |
| Hall, 2023 [15] | Band-pass filter | CNN | 86.23 |
| Reddy et al., 2023 [37] | Band-pass filter + DWT | DBN | 92.56 |
| Zhang et al., 2023 [61] | STFT + Dimensionality Reduction | CNN + LSTM | 95.74 |
| Wang et al., 2023 [85] | Band-pass filter | CNN + LSTM | 95 |
| Okreghe et al., 2023 [39] | PCA | Clustering | 91.53 |
| Saif-ur-Rehman et al., 2023, [40] | High-pass filter | CNN | 93 |
| Zacharelos et al., 2023 [52] | High-pass filter + Feature Extraction | ANN | 93.25 |
| Wang et al., 2023 [43] | PCA + Clustering | SVM, KNN | 96.1 |
| Li et al., 2023 [90] | Band-pass filter + PCA + WT | Deep RL | 98.4 |
| Sun et al., 2023 [62] | Band-pass filter + WT | CNN | 99.97 |
| Meyer et al., 2023 [91] | Band-pass filter | Running NN | 98.04 |
| Valencia and Alimohammad, 2023 [65] | Dimensionality Reduction | BNN | 93.88-93.90 |
| Meyer et al., 2023 [79] | High-pass filter | ANN | 89-100 |
| Liu et al., 2022 [86] | Band-pass filter + PCA | CNN + LSTM | 99 |
| Yi et al., 2022 [44] | Band-pass filter + Feature Extraction | CNN | 93.1 |
| Cheok, 2022 [88] | Feature Extraction | SNN | 93.31 |
| Wouters et al., 2021 [82] | High-pass filter + Feature Extraction | MLP | 96 |
| Saif-ur-Rehman et al., 2021 [41] | High-pass filter + PCA | CNN + Clustering | 92.3 |
| Radmanesh et al., 2021 [73] | Band-pass filter + Feature Extraction | Clustering | 90 |

| Eom et al., 2021 [76] | Band-pass filter + Feature Extraction | AE + Clustering | 99.81 |
|---|---|---|---|
| Sakamoto et al., 2021 [47] | Band-pass filter + Feature Extraction | KNN | 98.54 |
| Issar et al., 2020 [81] | Band-pass filter | ANN | - |
| Valencia and Alimohammad, 2020 [38] | Band-pass filter | BNN | 93 |
| Rácz et al., 2020 [45] | Band-pass filter | CNN | 89 |
| Li et al., 2020 [46] | Band-pass filter + PCA | CNN | 99.5 |
| Wouters et al., 2020 [83] | High-pass filter + Feature Extraction | ANN + Clustering | - |
| Park et al., 2019 [3] | Band-pass filter + PCA + Clustering | MLP | 96.42 |
| Bernert et al., 2019 [89] | Band-pass filter + Feature Extraction | SNN attention-based mechanism | 89 |
| Dai et al., 2019 [63] | Band-pass filter + Feature Extraction | SVM | 94.10-99.54 |
| Wu et al., 2019 [80] | Band-pass filter + Feature Extraction | GAN | 98.20 |
| Mukhopadhyay et al., 2018 [87] | Clustering | SNN | 96 |
| Noce et al., 2018 [69] | Feature Selection | SVM | 95.56 |
| Saif-ur-Rehman et al., 2018 [42] | Band-pass filter | CNN | 98.90 |
| Yang et al., 2017 [74] | High-pass filter | PCA + SVM | - |
| Yang et al., 2014 [68] | PCA + Feature Selection | DT | - |
| Ambard and Rotter, 2012 [72] | Dimensionality Reduction | SVM | - |
| Yang and Mason, 2011 [66] | Feature Extraction | Clustering | - |
| Adamos et al., 2010 [64] | Dimensionality Reduction + Clustering | ELM | - |
| Ding and Yuan, 2008 [51] | Band-pass filter + Clustering | SVM | - |
| Kim et al., 2006 [71] | Band-pass filter | SVM | - |
| Vogelstein et al., 2004 [75] | Band-pass filter + PCA | SVM | 95 |
| Chandra and Optican, 1997 [77] | Filtering + Clustering | ANN | - |
| Mirfakhraei and Horch, 1994 [78] | Band-pass filter | DNN | - |

***Appendix 3***. *Summary of datasets used in the spike classification studies.*

| *Authors* | *Type* | *Biological Source* | *Number of Channels* | *Recording Method* | *Sampling Rate (Hz)* | *Number of Classes* | *Purpose / Application* |
|---|---|---|---|---|---|---|---|
| Park et al. [3] | Simulated + In vivo | Mouse | N/A | Tetrode | 24,000 | 3 | To evaluate spike sorting algorithms on hybrid data |
| Sharifrazi et al. [13] | In vitro | Mosquito | 60 | MEA | 10,000 | 3 | Used to detect and classify evoked spike responses |
| Hall [15] | In vivo | Rhesus monkey | 96 (Utah) / 16 (Linear) | Utah / Linear array | 30,000 | 2 | Comparative spike classification across brain regions |
| Yang et al. [68] | In vivo | Not specified | Multiple | VLSI-based decision tree | 50,000 | Variable | Classify spikes using decision trees in real-time |
| Issar et al. [81] | In vivo | Rhesus monkey | 96 / 16 | Utah / Linear array | 30,000 | 2 | Assess classification under varying signal-to-noise conditions |
| Kim et al. [71] | Simulated | N/A | N/A | Simulated | N/A | N/A | Benchmark for spike sorting algorithms |
| Reddy et al. [37] | In vivo | Human neurons | 32 (FPGA) | Neural Signal Processor | Varied | Multiple | Real-time classification in a clinical setting |
| Valencia and Alimohammad [38] | In vivo | Human | MEA | BNN-based sorting | 24,000 | 3 | Human cortex recordings used for spike sorting |
| Zhang et al. [61] | Simulated | N/A | Wave_Clus | Log-Mel + CNN | Varied | Multiple | Synthetic dataset for feature learning and classification |
| Chandra and Optican [77] | In vivo | Monkey | Single-channel | Neural network | 24,000 | Multiunit | Spike classification using NN |
| | | | | | | | from single-channel data |
| Wang et al. [85] | Simulated + Experimental | Macaque monkey | 64 (MEA) | CNN + LSTM | 20–24,400 | 3–4 | Hybrid data for deep learning-based spike sorting |
| Ding and Yuan [51] | Simulated + Real | Chicken retina | 60 | MEA | 20,000 | 3 | Use of CNNs on retinal spike data |
| Wouters et al. [82] | In vivo | Not specified | High-density MEA | NN-based sorting | 30,000 | Multiple | Application of neural networks to large-channel recordings |
| Rácz et al. [45] | In vivo | Wistar rats | 128 (Silicon probe) | Deep learning-based | 20,000 | 20+ | Evaluates deep sorting on high-class-count data |
| Okreghe et al. [39] | In vivo | Human & NHPs | Varied | CNN-based artifact removal | Varied | Varied | Focus on artifact correction before spike classification |
| Saif-ur-Rehman et al. [40] | In vivo | Tetraplegic patients | 3072 | Self-organizing sorting | Varied | Adaptive | High-density patient data for adaptive sorting |
| Liu et al. [86] | Simulated | Macaque monkey | 64-site MEA | CNN + LSTM | 24,400 | Multiple | Evaluates deep temporal models on hybrid data |
| Yang and Mason [66] | Simulated, In vivo | Real neural recordings (used to generate synthetic signals) | N/A | Simulated spike trains from extracted templates | N/A | 5 | On-chip real neural data |
| Ambard and Rotter [72] | Simulated | N/A | N/A | SVM + PSP Kernels | Simulated | N/A | Uses PSP kernel features for SVM classification |
| Wang et al. [43] | In vivo | Cynomolgus Monkey | 128 | NN-based Pattern Recognition | Varied | Brain areas | Classification across brain regions using neural networks |

| Saif-ur-Rehman et al. [41] | In vivo | Human, NHP | 192 | Utah array, Micro-wires | 24,000 | 3 | Cross-species study using high-density probes |
| --- | --- | --- | --- | --- | --- | --- | --- |
| Li et al. [46] | In vivo | Macaque monkey | 128 | MEA | 24,400 | 3 | Standard benchmark for multi-class sorting in primates |
| Li et al. [90] | In vitro / In vivo | Macaque / Synthetic | N/A | MEA | 96,000 (downsampled) | N/A | Mixed-type dataset for evaluating generalizability |
| Adamos et al. [64] | In vitro | Rat / Beetle | 1 | ISOMAP, ELM | 24,000 | N/A | Feature extraction on rare and small-scale bio recordings |
| Radmanesh et al. [73] | Simulated | Synthetic | 1 | DCAE | 24,000 | 10 | Deep clustering autoencoder for spike types |
| Sun et al. [62] | In vivo | Mouse | N/A | SCN + CNN | N/A | 3 | Classification of SCN region activity using deep features |
| Eom et al. [76] | Simulated, In vivo | Mouse | 1, 4 | Tetrode | 30.3 kHz | N/A | Fusion of real and simulated data for model robustness |
| Lotfi and Daliri [84] | Simulated, In vivo | Monkey | 1, 4 | MEA | 24 kHz | N/A | Testing MEA data preprocessing pipelines |
| Meyer et al. [91] | Simulated, In vivo | Not specified | 1 | MEA | Not specified | N/A | Comparative dataset for general spike classification tasks |
| Mohammadi et al. [53] | Simulated, In vivo | Mouse (C57BL J) | 16 | MEA, Neuropixels probe | 30,000 | 8 | Synthetic + real data mix for algorithm testing |
| Yi et al. [44] | In vivo | Mouse | 49 | MEA, 64-channel electrode array | 30,000 | 27 | Multiple neural units from mouse cortex |
| Mukhopadhyay et al. [87] | Simulated | Rhesus Monkey | 16 | MEA | 8,000 | 3 | Spike classification benchmark |
| Valencia and Alimohammad [65] | Simulated | Rhesus Monkey | 20 | MEA | 40,000 | 3 | Used to compare classifiers under same conditions |
| Vogelstein et al. [75] | Simulated | Rhesus Monkey | 4 | MEA | 40,000 | 2 | Minimal class setting for simple classification |
| Yang et al. [74] | Simulated | Rhesus Monkey | 1 | MEA | 250 (high-pass filtered) | 3 | Simple configuration with low sampling |
| Cheok [88] | Simulated | N/A | N/A | N/A | 24,000 | 3 | Synthetic dataset for general testing |
| Sakamoto et al. [47] | In vivo | Mouse | 32 | Tetrode | 32,000 | Varies | Real spike data with varied classes |
| Meyer et al. [79] | Simulated | N/A | N/A | N/A | N/A | 9 | High-class synthetic spike dataset |
| Noce et al. [69] | Simulated | N/A | N/A | N/A | 20,000 | 10 | Used in deep learning classification |
| Hussein and Mohammed [70] | Simulated | N/A | N/A | N/A | 30,000 | 3 | Testing CNN-based classifiers |
| Bernert et al. [89] | Simulated | N/A | Multiple channels | Tetrode (possible) | N/A | N/A | Varied spike features mentioned |
| Dai et al. [63] | Simulated | N/A | Multiple channels | N/A | N/A | 3 types (S1, S2, S3) | Used to model different spike types |
| Wouters et al. [83] | Simulated | N/A | 10 | MEA | 30,000 | N/A | Neural network- |

| | | | | | | | based feature extraction |
|---|---|---|---|---|---|---|---|
| Saif-ur-Rehman et al. [42] | In vivo | Tetraplegic Patient | 192 | Utah Array | 30,000 | N/A | Human neural recordings for BCI applications |
| Wu et al. [80] | In vitro | Mouse | 8 | MEA Array | 20,000 | N/A | Cultured neuron data for spike sorting |